\documentclass{article}

\usepackage{microtype}
\usepackage{graphicx}
\usepackage{subfigure}
\usepackage{booktabs} 

\usepackage{hyperref}

\usepackage[accepted]{arxiv}

\usepackage{amsmath}
\usepackage{amssymb}
\usepackage{mathtools}
\usepackage{amsthm}

\usepackage[capitalize,noabbrev]{cleveref}

\theoremstyle{plain}
\newtheorem{theorem}{Theorem}[section]

\theoremstyle{definition}

\theoremstyle{remark}

\usepackage[textsize=tiny]{todonotes}

\usepackage{enumitem}
\DeclareMathOperator{\diag}{diag}
\DeclareMathOperator{\E}{\mathbb{E}}

\icmltitlerunning{MVG-CRPS: A Robust Loss Function for Multivariate Probabilistic Forecasting}

\begin{document}

\twocolumn[
\icmltitle{MVG-CRPS: A Robust Loss Function for Multivariate Probabilistic Forecasting}




\begin{icmlauthorlist}
\icmlauthor{Vincent Zhihao Zheng}{mcgill}
\icmlauthor{Lijun Sun}{mcgill}
\end{icmlauthorlist}

\icmlaffiliation{mcgill}{McGill University}

\icmlcorrespondingauthor{Lijun Sun}{lijun.sun@mcgill.ca}

\icmlkeywords{Probabilistic Forecasting, Scoring Rules, Time Series}

\vskip 0.3in
]



\printAffiliationsAndNotice{}  

\begin{abstract}
Multivariate Gaussian (MVG) distributions are central to modeling correlated continuous variables in probabilistic forecasting. Neural forecasting models typically parameterize the mean vector and covariance matrix of the distribution using neural networks, optimizing with the log-score (negative log-likelihood) as the loss function. However, the sensitivity of the log-score to outliers can lead to significant errors in the presence of anomalies. Drawing on the continuous ranked probability score (CRPS) for univariate distributions, we propose MVG-CRPS, a strictly proper scoring rule for MVG distributions. MVG‐CRPS admits a closed‐form expression in terms of neural network outputs, thereby integrating seamlessly into deep learning frameworks. Experiments on real-world datasets across multivariate autoregressive and univariate sequence-to-sequence (Seq2Seq) forecasting tasks show that MVG-CRPS improves robustness, accuracy, and uncertainty quantification in probabilistic forecasting.
\end{abstract}

\section{Introduction}
Probabilistic forecasting aims to capture the uncertainty inherent in time series data, offering a distribution of possible future values rather than single-point estimates provided by deterministic forecasts. This approach is invaluable in various domains such as finance \citep{groen2013real}, weather prediction \citep{palmer2012towards}, and health care management \citep{jones2012improved}, where understanding the range of potential outcomes is crucial for risk assessment and informed decision-making.

To achieve accurate probabilistic forecasts, it is crucial to develop a well-specified forecasting model and employ an effective evaluation metric during training. While the development of probabilistic forecasting models has received significant attention in recent years, the advancement of evaluation metrics has been less explored \citep{bjerregaard2021introduction}. A common approach for evaluating a forecaster involves applying scoring rules that assign a numerical score to each forecast, which are then collected across a test dataset. The choice of scoring rule typically depends on the type of forecasting task—whether it is deterministic or probabilistic, univariate or multivariate—and the nature of the variable being forecasted, such as categorical or continuous. In this paper, we focus on continuous variables within the probabilistic forecasting setting.

\begin{figure}[!t]
\centering\includegraphics[width=0.3\textwidth]{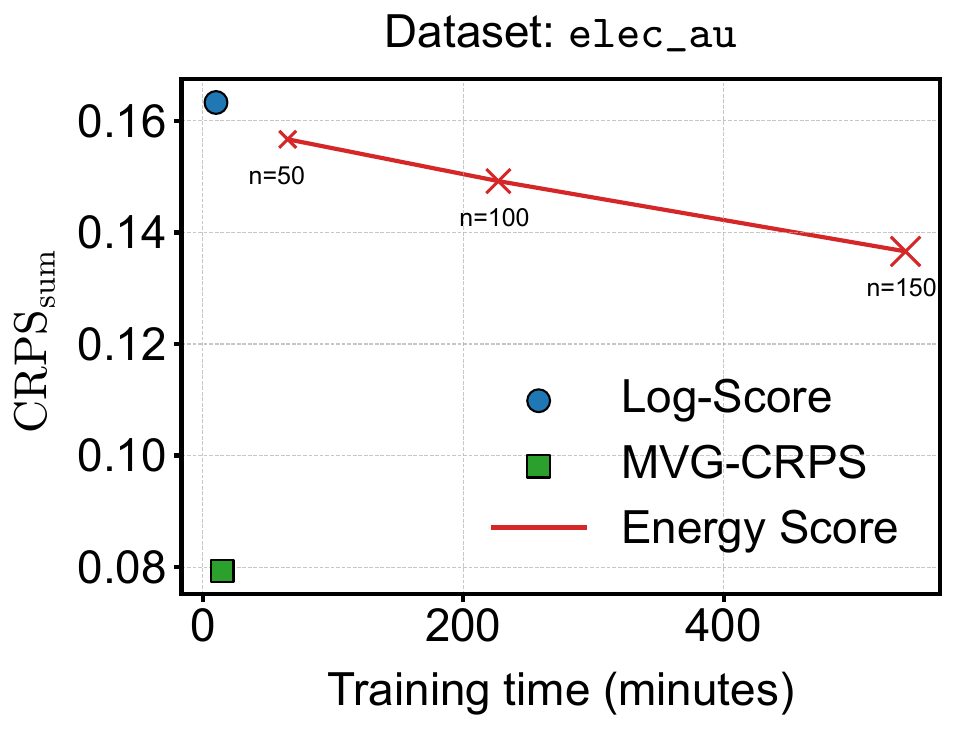}
  \caption{A motivating example illustrating how MVG-CRPS improves predictive performance by limiting outlier influence and reducing training time through its sampling-free design. The energy score is computed using sample sizes of 50, 100, and 150.}
\label{fig:compute_eff}
\end{figure}

Although scoring rules for univariate probabilistic forecasting are well-established, their multivariate counterparts remain less developed \citep{panagiotelis2023probabilistic}. Multivariate time series data, characterized by multiple interconnected variables, present additional challenges such as increased complexity, higher dimensionality, and intricate cross-variable interactions. Traditional evaluation metrics like the CRPS \citep{matheson1976scoring} and the log-score have been effective in univariate contexts but often fall short in multivariate scenarios due to issues like sensitivity to outliers and computational inefficiency. For example, when extended to the multivariate case, the CRPS lacks a closed-form expression, leading to computational inefficiency; additionally, the log-score heavily penalizes the tails of the data distribution, making it more sensitive to outliers \citep{gebetsberger2018estimation,bjerregaard2021introduction}.

To overcome these challenges, we introduce MVG-CRPS, a strictly proper scoring rule utilized as a loss function for multivariate probabilistic forecasting with a Gaussian distribution output layer. Our method utilizes a PCA whitening transformation to decorrelate the multivariate variables into a new random vector with zero mean and identity covariance matrix. Consequently, each component of this vector follows a standard Gaussian distribution, allowing us to apply the closed-form expression of the CRPS for evaluation. This approach not only overcomes the lack of a closed-form CRPS in the multivariate case but also reduces sensitivity to outliers and extreme tails, enhancing both robustness and computational efficiency. The advantages of our approach are illustrated in Fig.~\ref{fig:compute_eff}, where the model trained with MVG-CRPS achieves higher accuracy while significantly reducing training time. The key contributions of our work are:

\begin{itemize}[nosep, noitemsep]
    \item We propose a scoring rule for multivariate probabilistic forecasting that is less sensitive to outliers and extreme tails of the data distribution. Under the Gaussian assumption, we prove that this scoring rule is strictly proper. Its robustness ensures that the resulting forecaster prioritizes the overall data structure rather than being disproportionately affected by anomalies.
    \item The proposed scoring rule has a closed-form expression, allowing for the analytical computation of derivatives. This feature is particularly advantageous for training deep learning models, as it enables seamless integration with backpropagation algorithms and reduces computational overhead.
    \item We perform extensive experiments with deep probabilistic forecasting models on real-world datasets. Our results demonstrate that the proposed method balances accuracy and efficiency more effectively than common scoring rules.
\end{itemize}

\section{Related Work}

\subsection{Probabilistic Forecasting}
Probabilistic forecasting concentrates on modeling the probability distribution of target variables, as opposed to deterministic forecasting, which offers only single-point estimates. This approach is vital for capturing the uncertainty inherent in time series data, enabling better risk assessment and decision-making. There are two primary distributional regression methodologies for probabilistic forecasting: parametric (e.g., through probability density functions (PDFs)) and non-parametric (e.g., through quantile functions) \citep{benidis2022deep}.

Non-parametric approaches typically predict specific quantiles of the target distribution, avoiding the need for strict parametric assumptions. For example, MQ-RNN \citep{wen2017multi} generates quantile forecasts using a Seq2Seq RNN architecture. By forecasting multiple quantiles, these models provide an approximation of the target distribution, making them especially effective for capturing asymmetries or heavy-tailed behaviors. 

PDF-based methods involve assuming a specific probability distribution (e.g., Gaussian, Poisson) and using neural networks to estimate the parameters of that distribution. For example, DeepAR \citep{salinas2020deepar} utilizes a recurrent neural network (RNN) to model hidden state transitions and predict Gaussian distribution parameters at each time step. Its multivariate extension, GPVar \citep{salinas2019high}, employs a Gaussian copula to convert observations into Gaussian variables, thereby assuming a joint MVG distribution. This technique captures dependencies among multiple time series, which is crucial for accurate multivariate probabilistic forecasting. In the Gaussian case, one can further introduce a batch dimension and use generalized least squares (GLS) method to capture the autocorrelation and cross-correlation among different time steps \citep{zheng2024better,zheng2024multivariate}. 

Neural networks can also parameterize more complex probabilistic models, enabling greater flexibility and expressiveness. Examples include state space models (SSMs) \citep{rangapuram2018deep,de2020normalizing}, normalizing flows (NFs) \citep{rasul2020multivariate}, and diffusion models \citep{rasul2021autoregressive}. Copula-based methods have also been explored to model dependencies between multiple time series. Studies by \citet{drouin2022tactis} and \citet{ashok2023tactis} use copulas to construct multivariate distributions by specifying individual marginal distributions and a copula function that captures the dependence structure. This allows for flexible modeling of inter-variable relationships, which is essential in multivariate probabilistic forecasting. Among existing approaches that model the full probability distribution, the majority use the log-score as the loss function to optimize model parameters.

\subsection{Scoring Rules}

Scoring rules are fundamental tools used to evaluate the quality of probabilistic forecasts by assigning numerical scores based on the predicted probability distributions and the observed outcomes. A scoring rule is considered proper if the expected score is minimized when the predicted probability distribution \(p\) matches the true distribution \(q\) of the observations. Formally, a scoring rule \(s(p,q)\) is proper if the divergence \(d(p,q)=s(p,q)-s(q,q)\) is non-negative and it is strictly proper if \(d(p, q)=0\) implies \(p=q\) \citep{brocker2009reliability}.

One of the most widely used scoring rules in both univariate and multivariate settings is the log-score, also known as the negative log-likelihood (NLL). It is defined as the negative logarithm of the predictive density evaluated at the observed value. This score is particularly common when a parametric form of the predictive density is available \citep{panagiotelis2023probabilistic}. The log-score is a strictly proper scoring rule and has several desirable properties, such as consistency and sensitivity to the entire distribution. However, it is known to lack robustness because it heavily penalizes forecasts that assign low probability to the observed outcome, making it highly sensitive to outliers and extreme events \citep{gneiting2007probabilistic}.

To address the sensitivity of the log-score to outliers, CRPS has been proposed as a more robust alternative, especially in univariate settings \citep{rasp2018neural}. CRPS measures the difference between the predictive CDF (cumulative distribution function) and the observation, effectively integrating the absolute error over all possible threshold values. It can be interpreted as a generalized version of the mean absolute error (MAE) \citep{gneiting2005calibrated}. A key distinction between the log-score and CRPS is that CRPS grows linearly with the prediction error, while the log-score penalizes errors exponentially as the prediction deviates further from the observed value. Consequently, the log-score assigns harsher penalties to poor probabilistic forecasts, increasing its sensitivity to outliers.

CRPS has been effectively used for parameter estimation in probabilistic models \citep{gneiting2005calibrated,olivares2023probabilistic,lang2024aifs}. For example, the method of minimum CRPS estimation has been introduced for fitting ensemble model output statistics (EMOS) coefficients, optimizing the forecasts by directly minimizing the CRPS rather than maximizing the likelihood \citep{gneiting2005calibrated}. This approach can lead to better-calibrated predictive distributions that are not excessively wide or overdispersed.

Extending scoring rules to multivariate forecasting introduces additional challenges due to variable dependencies and increased dimensionality. Although the log-score can be applied in multivariate settings, its sensitivity to outliers persists. The energy score (ES) \citep{gneiting2007strictly}, a widely used multivariate extension of the CRPS, evaluates the discrepancy between predictive and observed distributions by measuring the expected distances between random vectors. While ES effectively detects errors in the forecast mean, it is less sensitive to variance errors and, more critically, to misspecifications in the correlation structure among variables \citep{pinson2013discrimination,alexander2024evaluating}. The absence of a closed form expression also necessitates the use of Monte Carlo simulations to approximate the score by drawing samples from the predictive distribution, which can be computationally intensive \citep{panagiotelis2023probabilistic}.

Other multivariate scoring rules include the Variogram Score \citep{scheuerer2015variogram} and the Dawid-Sebastiani Score \citep{wilks2020regularized}. These metrics also face computational challenges and may require approximations when closed-form solutions are unavailable. We refer readers to the comprehensive reviews by \citet{gneiting2014probabilistic}, \citet{ziel2019multivariate}, and \citet{tyralis2024review} for further details.

\section{Our Method}

\subsection{Multivariate Probabilistic Forecasting}

Probabilistic forecasting aims to estimate the joint distribution over a collection of future quantities based on a given history of observations \citep{gneiting2014probabilistic}. Denote the time series vector at a time point \(t\) as \(\mathbf{z}_{t}=\left[z_{1,t},\dots,z_{N,t}\right]^\top \in \mathbb{R}^{N}\), where $N$ is the number of series. The problem of probabilistic forecasting can be formulated as \(p\left(\mathbf{z}_{{T+1}:{T+Q}} \mid \mathbf{z}_{{T-P+1}:{T}}; \mathbf{x}_{{T-P+1}:{T+Q}}\right)\), where \(\mathbf{z}_{{t_1}:{t_2}} =\left[\mathbf{z}_{t_1},\ldots,\mathbf{z}_{t_2}\right]\), \(P\) is the conditioning range, \(Q\) is the prediction range, and \(T\) is the time point that splits the conditioning range and prediction range. \(\mathbf{x}_{t}\) are some known covariates for both past and future time steps. 

Multivariate probabilistic forecasting can be formulated in different ways. One way is over the time series dimension, where multiple interrelated variables are forecasted simultaneously at each time point. Considering an autoregressive model, where the predicted output is used as input for the next time step, this formulation can be factorized as
\begin{equation}\label{eqn:prob1}
\begin{aligned}
    &p\left(\mathbf{z}_{{T+1}:{T+Q}} \mid \mathbf{z}_{{T-P+1}:{T}}; \mathbf{x}_{{T-P+1}:{T+Q}}\right) \\ =&\prod_{t=T+1}^{T+Q} p\left(\mathbf{z}_{t} \mid \mathbf{z}_{{t-P}:{t-1}}; \mathbf{x}_{{t-P}:{t}}\right)  =\prod_{t=T+1}^{T+Q} p\left(\mathbf{z}_{t} \mid \mathbf{h}_{t}\right),
\end{aligned}
\end{equation}
where \(\mathbf{h}_t\) is a state vector that encodes all the conditioning information used to generate the distribution parameters, typically via a neural network.

\begin{figure}[!t]
  \centering\includegraphics[width=0.3\textwidth]{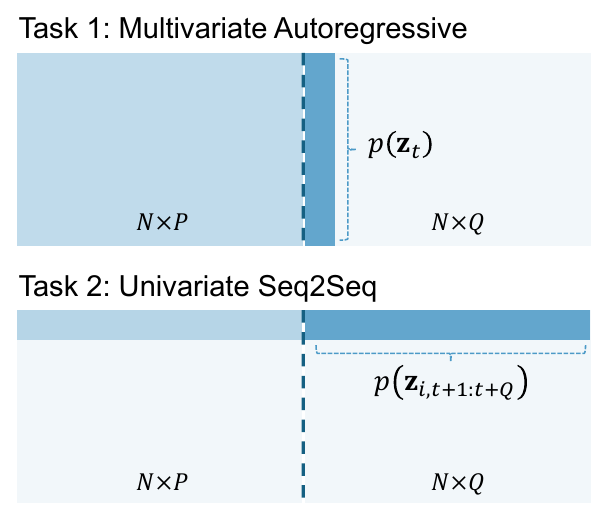}
  \caption{Illustration of the multivariate autoregressive and univariate Seq2Seq forecasting tasks.}
\label{fig:tasks}
\end{figure}

Another option is over the prediction horizon, where forecasts are made across multiple future time steps for one or more variables, capturing temporal dependencies and uncertainties over time. Considering a shared model across different series:
\begin{equation}\label{eqn:prob_uni_ss}
    p\left(\mathbf{z}_{i,{T+1}:{T+Q}} \mid \mathbf{z}_{i,{T-P+1}:{T}}; \mathbf{x}_{i,{T-P+1}:{T+Q}}\right),
\end{equation}
where \(i=1,\dots,N\) denotes the identifier of a particular time series. Since the model outputs forecasts for the entire prediction horizon directly, it is also called a Seq2Seq model. Without loss of generality, we use the first approach as an example to illustrate our method, since both approaches focus on estimating a multivariate distribution \(p\left(\mathbf{z}_{t}\right)\) or \(p\left(\mathbf{z}_{i,{T+1}:{T+Q}}\right)\) (Fig.~\ref{fig:tasks}).

A typical probabilistic forecasting model assumes Gaussian noise; for example, it models \(\mathbf{z}_t\) as jointly following a multivariate Gaussian distribution:
\begin{equation}
    \left.\mathbf{z}_{t} \mid \mathbf{h}_{t}\right.\sim\mathcal{N}\left( \boldsymbol{\mu}(\mathbf{h}_{t}), \boldsymbol{\Sigma}(\mathbf{h}_{t})\right),
\end{equation}
where \(\boldsymbol{\mu}(\cdot)\) and \(\boldsymbol{\Sigma}(\cdot)\) are the functions mapping \(\mathbf{h}_{t}\) to the mean and covariance parameters. The log-likelihood of the distribution given observed time series data up to time point \(T\) can be used as the loss function for optimizing a DL model: 
\begin{multline}\label{eqn:gls_ll}
    \mathcal{L}=\sum_{t=1}^{T} \log p\left(\mathbf{z}_{t} \mid \theta\left(\mathbf{h}_{t}\right)\right) \\
    \propto \sum_{t=1}^{T} -\frac{1}{2}[\ln \lvert \boldsymbol{\Sigma}_{t} \rvert + \boldsymbol{\eta}_{t}^\top\boldsymbol{\Sigma}_{t}^{-1}\boldsymbol{\eta}_{t}],
\end{multline}
where \(\boldsymbol{\eta}_{t}=\mathbf{z}_{t}-\boldsymbol{\mu}_{t}\). The above formulation simplifies to the univariate case when we set \(N=1\) for the model, with the same model being shared across all time series:
\begin{equation}\label{eqn:z_norm}
    \left.z_{i,t} \mid \mathbf{h}_{i,t}\right.\sim\mathcal{N}\left( \mu(\mathbf{h}_{i,t}), \sigma^2(\mathbf{h}_{i,t})\right),
\end{equation}
where $\mu(\cdot)$ and $\sigma(\cdot)$ map $\mathbf{h}_{i,t}$ to the mean and standard deviation of a Gaussian distribution. The corresponding log-likelihood becomes
\begin{equation}\label{eqn:nll_uni}
\mathcal{L}=\sum_{t=1}^{T} \sum_{i=1}^{N} \log p\left(z_{i,t} \mid \theta\left(\mathbf{h}_{i,t}\right)\right)\propto \sum_{t=1}^{T} \sum_{i=1}^{N} -\frac{1}{2}\epsilon_{i,t}^2-\ln \sigma_{i,t},
\end{equation}
where \(\epsilon_{i,t}=\frac{z_{i,t}-\mu_{i,t}}{\sigma_{i,t}}\). Eq.~\eqref{eqn:gls_ll} and Eq.~\eqref{eqn:nll_uni}, when used as scoring rules to optimize the model, are generally referred to as the log-score and are widely employed in probabilistic forecasting. 

For univariate problems, the CRPS is also a strictly proper scoring rule, defined as 
\begin{equation}\label{eqn:crps_uni_kernel}
    \operatorname{CRPS}(F, z)=\mathbb{E}_F|Z-z|-\frac{1}{2} \mathbb{E}_F\left|Z-Z^{\prime}\right|,
\end{equation}
where $F$ is the predictive CDF, $z$ is the observation, and $Z$ and $Z^{\prime}$ are independent random variables both associated with the CDF $F$. The CRPS has a closed-form expression when evaluating a Gaussian-distributed variable \(z \sim \mathcal{N}\left( \mu, \sigma^2\right)\) \citep{gneiting2005calibrated}:
\begin{equation}\label{eqn:crps_uni}
    \operatorname{CRPS}\left(\Phi, z\right)=z\left(2\Phi\left(z\right)-1\right)+2\varphi\left(z\right)-\frac{1}{\sqrt{\pi}},
\end{equation}
\begin{equation}\label{eqn:crps_uni_z}
    \operatorname{CRPS}\left(F_{\mu,\sigma}, z\right)=\sigma \operatorname{CRPS}\left(\Phi, \frac{z-\mu}{\sigma}\right),
\end{equation}
where \(F_{\mu,\sigma}\left(z\right)=\Phi\left(\frac{z-\mu}{\sigma}\right)\), \(\Phi\) and \(\varphi\) are the CDF and PDF of the standard Gaussian distribution.

The CRPS has been shown to be a more robust alternative to the log-score as a loss function in many problems \citep{gneiting2005calibrated,rasp2018neural,murad2021probabilistic,chen2024generative}. We observe that the log-score can grow arbitrarily large in magnitude when a single outlier disproportionately influences the loss function, owing to the unbounded nature of the logarithmic function (Eq.~\eqref{eqn:gls_ll} and Eq.~\eqref{eqn:nll_uni}). Additionally, the quadratic form of the error terms in the Gaussian likelihood also makes it sensitive to outliers (e.g., \(\epsilon_{i,t}^2\) in Eq.~\eqref{eqn:nll_uni}). In contrast, CRPS evaluates the entire predictive distribution rather than concentrating solely on the likelihood of individual data points (Eq.~\eqref{eqn:crps_uni}). Moreover, the CRPS can directly replace the log-score, providing analytical gradients with respect to \(\mu\) and \(\sigma\) for backpropagation. However, for a MVG distribution, the CRPS does not have a widely used closed-form expression.

\subsection{MVG-CRPS as Loss Function for Multivariate Forecasting}
In multivariate probabilistic forecasting, proper scoring rules such as the log-score (Eq.~\eqref{eqn:gls_ll}) and the energy score (\(\operatorname{ES}\)) are used to evaluate predictive performance. The energy score generalizes CRPS to assess probabilistic forecasts of vector-valued random variables \citep{gneiting2007strictly}:
\begin{equation}\label{eqn:es}
    \operatorname{ES}(F, \mathbf{z}) = \mathop{\E}_{\boldsymbol{Z}\sim F} \lVert \boldsymbol{Z} - \mathbf{z} \rVert^{\beta} - \frac{1}{2} \mathop{\E}_{\substack{\boldsymbol{Z}\sim F \\ \boldsymbol{Z}' \sim F}} \lVert \boldsymbol{Z} - \boldsymbol{Z}' \rVert^{\beta},
\end{equation}
where \(\lVert \cdot \rVert\) denotes the Euclidean norm and \(\beta = 1\) is commonly used in the literature \citep{ashok2023tactis}. With \(\beta = 1\), the energy score essentially becomees a multivariate extension of CRPS and grows linearly with respect to the norm, making it less sensitive to outliers compared to the log-score. Since there is no simple closed-form expression for Eq.~\eqref{eqn:es}, it is often approximated using Monte Carlo methods, where multiple samples are drawn from the forecast distribution to approximate the expected values:
\begin{equation}\label{eqn:es_approx}
    \operatorname{ES}(F, \mathbf{z}) = \frac{1}{n}\sum_{i=1}^{n} \lVert \boldsymbol{Z}_i - \mathbf{z} \rVert^{\beta} -  \frac{1}{2n^2}\sum_{i=1}^{n}\sum_{j=1}^{n} \lVert \boldsymbol{Z}_i - \boldsymbol{Z}_j \rVert^{\beta}.
\end{equation}
However, a significant disadvantage of using Eq.~\eqref{eqn:es_approx}) as the loss function is that it requires constant sampling during the training process, which can substantially slow down training.

In this section, we propose MVG-CRPS, a robust and efficient loss function designed as an alternative for multivariate forecasting. This loss function grows linearly with the prediction error, making it more robust than the log-score. Additionally, it does not require sampling during the training process, rendering it more efficient than the energy score.

Our proposed method is based on the whitening transformation of a time series vector that follows a MVG distribution, \(\mathbf{z}_{t} \sim \mathcal{N}\left(\boldsymbol{\mu}_{t}, \boldsymbol{\Sigma}_t\right)\). The whitening process transforms a random vector with a known covariance matrix into a new random vector whose covariance matrix is the identity matrix. As a result, the elements of the transformed vector have unit variance and are uncorrelated. This transformation begins by performing the singular value decomposition (SVD) of the covariance matrix:
\begin{equation}
\boldsymbol{\Sigma}_t = \boldsymbol{U}_t \boldsymbol{S}_t \boldsymbol{U}_t^{\top},
\end{equation}
where \(\boldsymbol{S}_t=\operatorname{diag}(\lambda_{i,t})_{i=1}^N\) is a diagonal matrix containing the eigenvalues \(\lambda_{i,t}\) of \(\boldsymbol{\Sigma}_t\), and \(\boldsymbol{U}_t\) is the orthonormal matrix of corresponding eigenvectors. We then define
\begin{equation}
\mathbf{v}_{t} = \boldsymbol{U}_t^{\top} \left( \mathbf{z}_{t} - \boldsymbol{\mu}_{t} \right),
\end{equation}
where \(\mathbf{v}_{t}\sim\mathcal{N}\left(\boldsymbol{0},\boldsymbol{S}_t\right)\) is a random vector with a decorrelated MVG distribution, having variances \(\lambda_i\) (i.e., the corresponding eigenvalue) along the diagonal of its covariance matrix. Next, we define
\begin{equation}
\mathbf{w}_{t} = \boldsymbol{S}_t^{-\frac{1}{2}} \mathbf{v}_{t} = \boldsymbol{S}_t^{-\frac{1}{2}} \boldsymbol{U}_t^{\top} \left( \mathbf{z}_{t} - \boldsymbol{\mu}_{t} \right),
\end{equation}
where \(\mathbf{w}_{t}\) is a random vector with each element following a standard Gaussian distribution, i.e., \({w}_{i,t} \sim \mathcal{N}\left(0, 1\right)\). We can then apply Eq.~\eqref{eqn:crps_uni} individually to each element and formulate the MVG-CRPS mimicking Eq.~\eqref{eqn:crps_uni_z} for multivariate problem:
\begin{equation}
\mathcal{L}\left(\mathcal{N}\left(\boldsymbol{\mu}_t, \boldsymbol{\Sigma}_t\right),\mathbf{z}_t\right)= \sum_{i=1}^N \sqrt{\lambda_{i,t}}\operatorname{CRPS}\left(\Phi,{w}_{i,t}\right). 
\end{equation}
The overall loss function for training the model is then formulated over an observation period \(T\):
\begin{equation}
\mathcal{L} = \sum_{t=1}^{T} \mathcal{L}\left(\mathcal{N}\left(\boldsymbol{\mu}_t, \boldsymbol{\Sigma}_t\right),\mathbf{z}_t\right). 
\end{equation}

The primary advantage of this loss function is its ability to exploit the closed-form expression of \(\operatorname{CRPS}\) by decorrelating time series variables through whitening. Applied to each transformed (whitened) variable, it evaluates the marginal distributions in the transformed space, where the whitening process is determined by the original covariance matrix. Consequently, the optimization process retains information about the covariance structure of the original distribution. Under the Gaussian assumption, MVG-CRPS is a strictly proper scoring rule (see Appendix \S \ref{apx:proof}).

\section{Experiments}
\subsection{Datasets and Models}

We apply the MVG-CRPS to two multivariate forecasting tasks: multivariate autoregressive forecasting and univariate Seq2Seq forecasting. For the first task, we employ two benchmark models: the RNN-based GPVar \citep{salinas2019high} and a decoder-only Transformer \citep{radford2018improving}. For the second task, we use the N-HiTS model, which is based on multi-layer perceptrons (MLPs) \citep{challu2023nhits}.

To generate the distribution parameters for probabilistic forecasting, we employ a Gaussian distribution head based on the hidden state \(\mathbf{h}_{i,t}\) produced by the model. Specifically, for the multivariate autoregressive forecasting, following \citet{salinas2019high}, we parameterize the mean vector as \(\boldsymbol{\mu}\left(\mathbf{h}_{t}\right)=\left[\mu_1\left(\mathbf{h}_{1, t}\right),\dots,\mu_N\left(\mathbf{h}_{N, t}\right)\right]^\top \in \mathbb{R}^N\) and 
adopt a low-rank-plus-diagonal parameterization of the covariance matrix \(\boldsymbol{\Sigma}\left(\mathbf{h}_t\right)=\boldsymbol{L}_{t}\boldsymbol{L}_{t}^\top+\diag{(\mathbf{d}_{t})}\), where $\mathbf{d}_t=[d_1\left(\mathbf{h}_{1, t}\right),\ldots,d_N\left(\mathbf{h}_{N, t}\right)]^{\top} \in \mathbb{R}_{+}^N$ and $\boldsymbol{L}_{t}=\left[\mathbf{l}_1\left(\mathbf{h}_{1, t}\right) ,\dots,\mathbf{l}_N\left(\mathbf{h}_{N, t}\right) \right]^\top \in \mathbb{R}^{N\times R}$, \(R \ll N\) is the rank parameter. Here, \(\mu_{i}(\cdot)\), \(d_{i}(\cdot)\), and \(\mathbf{l}_{i}(\cdot)\) are the mapping functions that generate the mean and covariance parameters for each time series \(i\) based on the hidden state \(\mathbf{h}_{i={1:N},t}\). In practice, we use shared mapping functions across all time series, denoted as \(\mu_{i} = \tilde{\mu}\), \(d_{i} = \tilde{d}\), and \(\mathbf{l}_{i} = \tilde{\mathbf{l}}\). This parameterization ensures that $\boldsymbol{\Sigma}(\mathbf{h}_t)$ is positive definite and efficiently parameterized. The diagonal component provides stability, while the low-rank component captures the covariance structure. The Gaussian assumption also enables the use of random subsets of time series (i.e., batch size \(B \leq N\)) for model optimization in each iteration, making it feasible to apply our method to high-dimensional time series datasets. Similarly, in the univariate Seq2Seq forecasting task, the mean \(\boldsymbol{\mu}\left(\mathbf{h}_{i}\right)\) and covariance \(\boldsymbol{\Sigma}(\mathbf{h}_{i})\) are defined over the forecast horizon for each specific time series, based on the hidden states \(\mathbf{h}_{i,t=T+1:T+Q}\). As a result, we can model the joint distribution \(p\left(\mathbf{z}_{i,{T+1}:{T+Q}}\right)\) over the forecasted values.

We implemented our models using PyTorch Forecasting \citep{pytorchforecasting}, with input data consisting of lagged time series values and covariates. We conducted extensive experiments on diverse real-world time series datasets obtained from GluonTS \citep{alexandrov2020gluonts} (see Appendix \S \ref{apx:data}). The prediction horizon ($Q$) and the number of rolling evaluations were taken from the configuration within GluonTS, where we follow the default setting by equating the context range to the prediction range, i.e., $P = Q$. Each dataset was sequentially split into training, validation, and testing sets. Each time series was individually normalized using a scaler fitted to its own training data \citep{salinas2020deepar,kim2021reversible}. Predictions were then rescaled to their original values for computing evaluation metrics. For comprehensive details on the experimental setup, we refer readers to Appendix \S \ref{apx:exp}.

\subsection{Toy Example}

\begin{figure}[!t]  \centering
\includegraphics[width=0.47\textwidth]{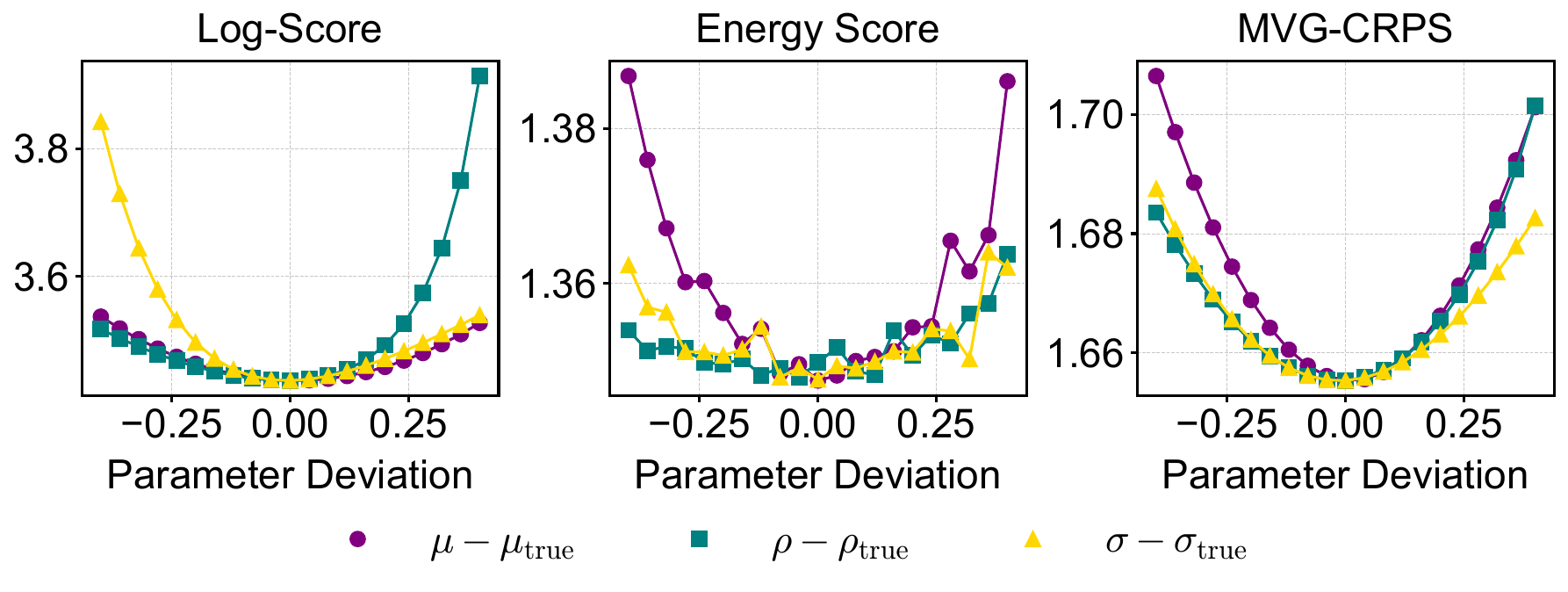}
\caption{Sensitivity of scoring rules to parameter deviations in the predicted mean, standard deviation, and correlation coefficient from the true data distribution (\(\mu_{\text{true}} = 1, \sigma_{\text{true}} = 1, \rho_{\text{true}} = 0.4)\). The energy score is computed with a sample size of 500.}
\label{fig:score_sens}
\end{figure}

We first perform a toy experiment following \citet{roordink2023scoring} using a true distribution $P = \mathcal{N}\left( 
\begin{bmatrix} 
1 \\ 
-1 
\end{bmatrix}, 
\begin{bmatrix} 
1 & 0.8 \\ 
0.8 & 4 
\end{bmatrix} 
\right)
$ and a predictive distribution $Q = \mathcal{N}\left( 
\begin{bmatrix} 
\mu \\ 
-1 
\end{bmatrix}, 
\begin{bmatrix} 
\sigma^2 & 2\rho\sigma \\ 
2\rho\sigma & 4 
\end{bmatrix} 
\right)
$, where we control the deviation of the three parameters $\mu, \rho, \sigma$ to study the various properties of different scores. As shown in Fig.~\ref{fig:score_sens}, the log-score exhibits high sensitivity to parameter deviations, particularly to inaccuracies in standard deviation ($\sigma$) and correlation coefficient ($\rho$) estimation. While the energy score is less sensitive than the log-score, it can occasionally produce slightly inconsistent scores due to its sample-based approximation (ideally, larger deviations result in higher scores). Furthermore, the energy score demonstrates lower sensitivity to deviations in correlation. In contrast, the proposed MVG-CRPS provides a well-balanced sensitivity across all three parameters while maintaining consistent scoring patterns, owing to its closed-form expression. Additionally, the strict properness of MVG-CRPS under the Gaussian assumption implies that the score is uniquely minimized when $P=Q$.

\subsection{Quantitative Evaluation}

We evaluate the MVG-CRPS by comparing it with models trained using the log-score and energy score. The evaluation metrics are \(\operatorname{CRPS}_{\text{sum}}\) and energy score (Eq.~\eqref{eqn:es_approx}). \(\operatorname{CRPS}_{\text{sum}}\) is computed by summing both the forecast and ground-truth values across all time series and then calculating the \(\operatorname{CRPS}\) (Eq.~\eqref{eqn:crps_uni_kernel}) over the resulting sums \citep{salinas2019high,drouin2022tactis,ashok2023tactis}: 
\begin{equation}
    \operatorname{CRPS}_{\text{sum}}=\E_{t}\left[\operatorname{CRPS}\left(F_{t}, \sum_iz_{i,t}\right)\right],
\end{equation}
where the empirical \(F_{t}\) is calculated by aggregating samples across time series, with 100 samples drawn to compute \(\operatorname{CRPS}_{\text{sum}}\). The evaluation using the energy score is provided in the Appendix \S \ref{apx:additional_results}.

\begin{table*}[!t]
\setlength{\tabcolsep}{2pt}
\scriptsize
\caption{Comparison of \(\operatorname{CRPS}_{\text{sum}}\) across different scoring rules in the multivariate autoregressive forecasting task. The best scores are in boldface. MVG-CRPS scores are underlined when they are not the best overall but exceed the log-score. Mean and standard deviation are reported from 10 runs of each model.}
\label{tab:crps_sum}
\begin{center}
\begin{tabular}{lccccccc}
\toprule
                          & VAR     & \multicolumn{3}{c}{GPVar}              & \multicolumn{3}{c}{Transformer} \\
\cmidrule(lr){3-8}
                          &       & log-score      & energy score        & MVG-CRPS    & log-score       & energy score              & MVG-CRPS   \\
\midrule
$\mathtt{elec\_au}$  & N/A   & 0.1261$\pm$0.0009   &   \textbf{0.0887$\pm$0.0004}     & \underline{0.0967$\pm$0.0008}  & 0.1633$\pm$0.0005    &    0.1492$\pm$0.0006   & \textbf{0.0793$\pm$0.0004}  \\
$\mathtt{cif\_2016}$       & 1.0000$\pm$0.0000   & 0.0122$\pm$0.0004   &   0.0420$\pm$0.0006    & \textbf{0.0111$\pm$0.0005}  & 0.0118$\pm$0.0003    &    0.0240$\pm$0.0014    & \textbf{0.0107$\pm$0.0002}  \\
$\mathtt{electricity}$            & 0.1315$\pm$0.0006   & 0.0419$\pm$0.0008   &   0.0616$\pm$0.0004    & \textbf{0.0249$\pm$0.0006}  & 0.0362$\pm$0.0002    &    0.0368$\pm$0.0004   & \textbf{0.0294$\pm$0.0004}  \\
$\mathtt{elec\_weekly}$              & 0.1126$\pm$0.0011   & 0.1515$\pm$0.0028   &   \textbf{0.0417$\pm$0.0014}     & \underline{0.0772$\pm$0.0031}  & 0.0937$\pm$0.0026    &    \textbf{0.0403$\pm$0.0013}   & \underline{0.0448$\pm$0.0014}  \\
$\mathtt{exchange\_rate}$      & 0.0033$\pm$0.0000   & 0.0207$\pm$0.0004   &   \textbf{0.0030$\pm$0.0001}     & \underline{0.0041$\pm$0.0001}  & \textbf{0.0047$\pm$0.0003} &   0.0067$\pm$0.0003  & 0.0091$\pm$0.0004           \\
$\mathtt{kdd\_cup}$ & N/A   & 0.3743$\pm$0.0019   &   0.3210$\pm$0.0019     & \textbf{0.2358$\pm$0.0014}  & 0.2076$\pm$0.0013      &   0.4789$\pm$0.0030  & \textbf{0.1959$\pm$0.0017}  \\
$\mathtt{m1\_yearly}$                       & N/A   & 0.4397$\pm$0.0041   &   0.4801$\pm$0.0022     & \textbf{0.3566$\pm$0.0029}  & 0.5344$\pm$0.0109       &   \textbf{0.3291$\pm$0.0047}  & \underline{0.4563$\pm$0.0111}  \\
$\mathtt{m3\_yearly}$                       & N/A   & 0.3607$\pm$0.0084   &   0.2186$\pm$0.0042     & \textbf{0.1423$\pm$0.0053}  & 0.3156$\pm$0.0102       &   0.4050$\pm$0.0061  & \textbf{0.2325$\pm$0.0094}  \\
$\mathtt{nn5\_daily}$     & 0.2303$\pm$0.0005   & 0.0998$\pm$0.0004   &   0.0958$\pm$0.0003     & \textbf{0.0948$\pm$0.0003}  & 0.0991$\pm$0.0003      &   0.0883$\pm$0.0004  & \textbf{0.0811$\pm$0.0002}  \\
$\mathtt{saugeenday}$                       & N/A   & 0.4040$\pm$0.0047   &   \textbf{0.3733$\pm$0.0048}     & \underline{0.3941$\pm$0.0055}  & 0.3771$\pm$0.0088      &   \textbf{0.3689$\pm$0.0053}  & \underline{0.3705$\pm$0.0047}  \\
$\mathtt{sunspot}$        & N/A   & 18.7115$\pm$1.3296  &   23.3988$\pm$0.9662     & \textbf{17.2438$\pm$0.5833} & 39.7454$\pm$1.4841    &    \textbf{16.6556$\pm$0.6167}   & \underline{22.6495$\pm$0.6752} \\
$\mathtt{tourism}$               & 0.1394$\pm$0.0012   & 0.2217$\pm$0.0027   &   0.2112$\pm$0.0014     & \textbf{0.2004$\pm$0.0022}  & 0.2100$\pm$0.0017     &    0.2087$\pm$0.0020   & \textbf{0.2082$\pm$0.0015}  \\
$\mathtt{traffic}$                     & 3.5241$\pm$0.0084   & 0.0742$\pm$0.0004 &  \textbf{0.0505$\pm$0.0002}
 & 0.0868$\pm$0.0002           & \textbf{0.0658$\pm$0.0002} &  0.0667$\pm$0.0002 & 0.0683$\pm$0.0000     \\
\bottomrule
\end{tabular}
\end{center}
\end{table*}

\begin{table}[!t]
\setlength{\tabcolsep}{1pt}
\scriptsize
\caption{Comparison of training cost (in minutes) for GPVar under different scoring rules in the multivariate autoregressive forecasting task.}
\label{tab:training_time}
\begin{center}
\begin{tabular}{lcccccc}
\toprule
                                              & \multicolumn{2}{c}{log-score}    & \multicolumn{2}{c}{energy score}     & \multicolumn{2}{c}{MVG-CRPS}   \\
\cmidrule(lr){2-7}
                                              & per epoch   & total   & per epoch   & total       & per epoch    & total   \\
\midrule
$\mathtt{elec\_au}$    & 0.86         & 33.53             & 16.29        & 717                   & \textbf{0.78}          & \textbf{29.14}        \\
$\mathtt{cif\_2016}$                          & 0.13         & \textbf{1.577}             & 4.83         & 401.04                & \textbf{0.12}          & 3.85         \\
$\mathtt{electricity}$                  & 0.40         & 67.38             & 11.17        & 782.4                 & \textbf{0.38}          & \textbf{22.7}         \\
$\mathtt{elec\_weekly}$                & 0.30         & \textbf{14.61}             & 10.95        & 383.52                & \textbf{0.26}          & 18.77        \\
$\mathtt{exchange\_rate}$               & \textbf{0.25}         & \textbf{16.4}              & 10.20        & 663.6                 & 0.29          & 23.63        \\
$\mathtt{kdd\_cup}$   & \textbf{0.42}         & \textbf{11.32}             & 14.23        & 2063.52               & \textbf{0.42}          & 28.79        \\
$\mathtt{m1\_yearly}$                         & 0.19         & \textbf{3.708}             & 5.66         & 469.92                & \textbf{0.18}          & 8.019        \\
$\mathtt{m3\_yearly}$                         & 0.43         & \textbf{7.299}             & 10.80        & 291.72                & \textbf{0.42}          & 14.49        \\
$\mathtt{nn5\_daily}$       & 0.29         & \textbf{9.208}             & 11.64        & 244.5                 & \textbf{0.27}          & 14.53        \\
$\mathtt{saugeenday}$                         & 0.23         & \textbf{12.65}             & 10.70        & 524.46                & \textbf{0.15}          & 15.32        \\
$\mathtt{sunspot}$          & 0.44         & 26.85             & 10.73        & 397.26                & \textbf{0.42}          & \textbf{16.96}        \\
$\mathtt{tourism}$                 & 0.49         & 23.96             & 10.56        & 243                   & \textbf{0.46}          & \textbf{12.51}        \\
$\mathtt{traffic}$                      & 0.94         & \textbf{76.98}             & 14.92        & 1044.6                & \textbf{0.92}          & 92.46        \\
\bottomrule
\end{tabular}
\end{center}
\end{table}

\begin{table}[!t]
\setlength{\tabcolsep}{2pt}
\scriptsize
\caption{Comparison of $\operatorname{CRPS}_{\text{sum}}$ across different scoring rules in the univariate Seq2Seq forecasting task.}
\label{tab:crps_sum_ss}
\begin{center}
\begin{tabular}{lccc}
\toprule
                               & \multicolumn{3}{c}{N-HiTS}            \\
\cmidrule(lr){2-4}
                               & log-score        & energy score        & MVG-CRPS    \\
\midrule
$\mathtt{covid}$          & 0.1297$\pm$0.0048 & N/A                        & \textbf{0.1011$\pm$0.0022} \\
$\mathtt{elec\_hourly}$   & 0.0470$\pm$0.0008 & N/A                        & \textbf{0.0398$\pm$0.0004} \\
$\mathtt{electricity}$           & 0.0409$\pm$0.0003 & 0.0378$\pm$0.0006          & \textbf{0.0372$\pm$0.0003} \\
$\mathtt{exchange\_rate}$ & 0.0089$\pm$0.0005 & 0.0060$\pm$0.0002          & \textbf{0.0053$\pm$0.0002} \\
$\mathtt{m4\_hourly}$     & 0.0649$\pm$0.0007 & 0.0595$\pm$0.0005          & \textbf{0.0399$\pm$0.0007} \\
$\mathtt{nn5\_daily}$     & 0.0571$\pm$0.0003 & 0.0876$\pm$0.0006          & \textbf{0.0569$\pm$0.0004} \\
$\mathtt{pedestrian}$     & 0.7985$\pm$0.0511 & 0.9110$\pm$0.0210          & \textbf{0.5296$\pm$0.0071} \\
$\mathtt{saugeenday}$     & 0.4804$\pm$0.0150 & 0.4372$\pm$0.0100          & \textbf{0.3864$\pm$0.0035} \\
$\mathtt{taxi\_30min}$    & 0.0496$\pm$0.0002 & 0.0603$\pm$0.0002          & \textbf{0.0449$\pm$0.0001} \\
$\mathtt{traffic}$        & 0.2065$\pm$0.0007 & \textbf{0.0815$\pm$0.0001} & \underline{0.0832$\pm$0.0002}    \\
$\mathtt{uber\_hourly}$   & 0.7027$\pm$0.0209 & 0.6461$\pm$0.0052          & \textbf{0.5380$\pm$0.0033} \\
$\mathtt{wiki}$           & 0.0660$\pm$0.0011 & \textbf{0.0429$\pm$0.0003} & \underline{0.0465$\pm$0.0004}     \\
\bottomrule
\end{tabular}
\end{center}
\end{table}

Table~\ref{tab:crps_sum} presents a comparison of \(\operatorname{CRPS}_{\text{sum}}\) across different scoring rules in the multivariate autoregressive forecasting task, underscoring the effectiveness of the  MVG-CRPS. MVG-CRPS consistently outperforms the log-score across most datasets. For instance, on the $\mathtt{electricity}$ and $\mathtt{tourism}$ datasets, MVG-CRPS achieves scores of 0.0249 and 0.2004, respectively, outperforming the scores of 0.0419 and 0.2217 achieved by the log-score. This indicates that MVG-CRPS leads to models that produce higher-quality forecasts. As discussed in later sections, this improvement is attributed to MVG-CRPS being less sensitive to outliers.

When compared to the energy score, MVG-CRPS demonstrates superior performance in 8 out of 13 datasets when using GPVar, and in 7 out of 13 datasets when using the Transformer model. For the remaining datasets, MVG-CRPS typically outperforms the log-score while maintaining comparable performance to the energy score, as highlighted in Table~\ref{tab:crps_sum} (underlined). Additionally, as shown in Table~\ref{tab:training_time}, the energy score requires significantly more training time compared to both the log-score and MVG-CRPS. Although MVG-CRPS has a slightly longer overall training time than the log-score, it is faster per epoch, suggesting that it may require more epochs to converge but is computationally efficient per iteration.

The results for the univariate Seq2Seq forecasting task are presented in Table~\ref{tab:crps_sum_ss} and align with the findings from the multivariate autoregressive task. Overall, these findings demonstrate that MVG-CRPS provides a more accurate alternative to the log-score and a more efficient one compared to the energy score.

\subsection{Qualitative Evaluation}

To further demonstrate the robustness and effectiveness of the MVG-CRPS loss, we compare the output covariance matrices from models trained with both the log-score and MVG-CRPS, and visualize the corresponding probabilistic forecasts.

\subsubsection{Multivariate Autoregressive Forecasting}

\begin{figure}[!t]
  \centering\includegraphics[width=0.46\textwidth]{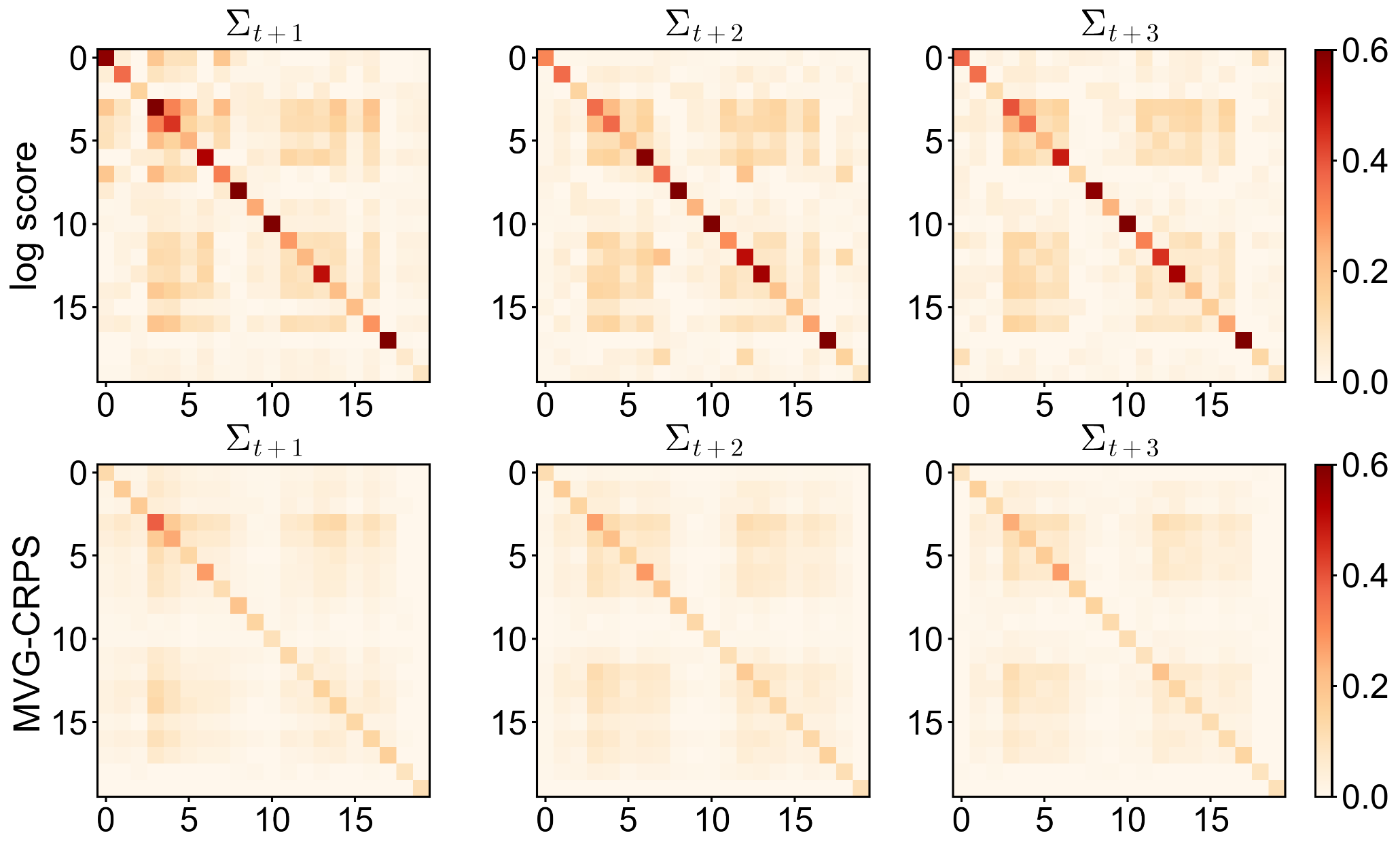}
  \caption{Comparison of output covariance matrices \(\boldsymbol{\Sigma}_{t}\) from GPVar on the \(\mathtt{elec\_weekly}\) dataset. The top and bottom rows display covariance matrices from models trained using the log-score and MVG-CRPS. For visual clarity, covariance values are clipped between 0 and 0.6.}
\label{fig:cov_b}
\end{figure}

We first examine the multivariate autoregressive forecasting task using the $\mathtt{elec\_weekly}$ dataset (Fig.~\ref{fig:cov_b}). When the model is trained with the log-score, the resulting covariance matrices exhibit disproportional variance values and occasionally display large covariance values. This observation is counterintuitive, given that the data have been normalized using scalers fitted to each time series. Such behavior can arise when large errors in the tails of the data disproportionately influence model training, causing the model to produce larger variances and covariances to accommodate these errors. In contrast, the model trained with MVG-CRPS produces covariance matrices with a more balanced distribution of variances and covariances. This indicates that MVG-CRPS effectively mitigates the influence of outliers, resulting in more reliable estimations of the predictive distribution.

\begin{figure}[!t]
  \centering\includegraphics[width=0.46\textwidth]{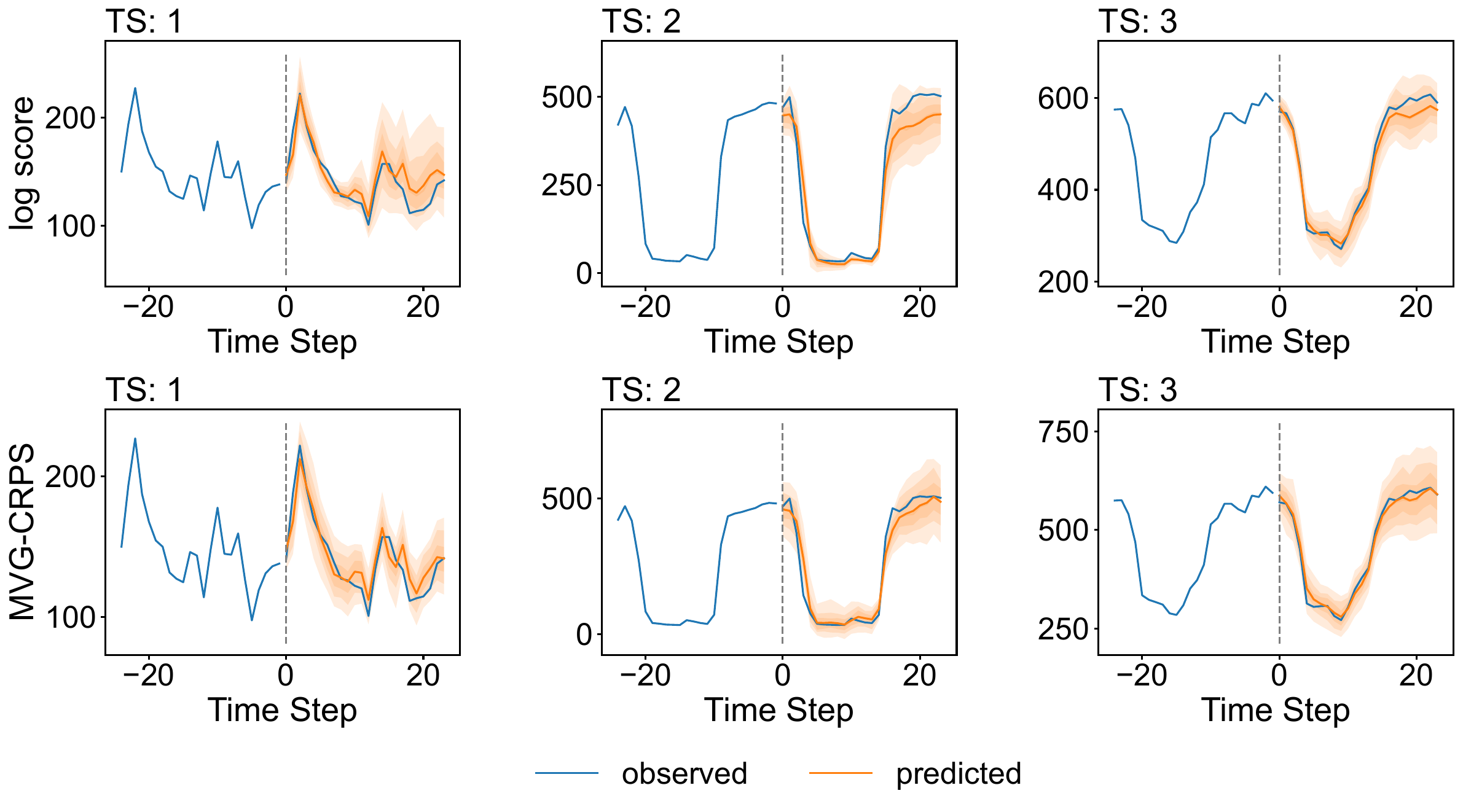}
  \caption{Comparison of probabilistic forecasts from GPVar on the $\mathtt{electricity}$ dataset. The first and second rows display forecasts from models trained using the log-score and MVG-CRPS.}
\label{fig:frcs_compare_b}
\end{figure}

To further illustrate the practical implications of our observations, we compare the probabilistic forecasts generated by GPVar on the $\mathtt{electricity}$ dataset (Fig.~\ref{fig:frcs_compare_b}). The forecasts from the MVG-CRPS-trained model are noticeably more calibrated and sharper, especially when the predictions are projected to more reliable outcomes. In contrast, the model trained with the log-score occasionally produces much wider prediction intervals than MVG-CRPS, highlighting its susceptibility to outliers and resulting in less reliable forecasts (e.g., TS 1 in Fig.~\ref{fig:frcs_compare_b}).


\subsubsection{Univariate Seq2Seq Forecasting}

\begin{figure}[!t]
  \centering\includegraphics[width=0.46\textwidth]{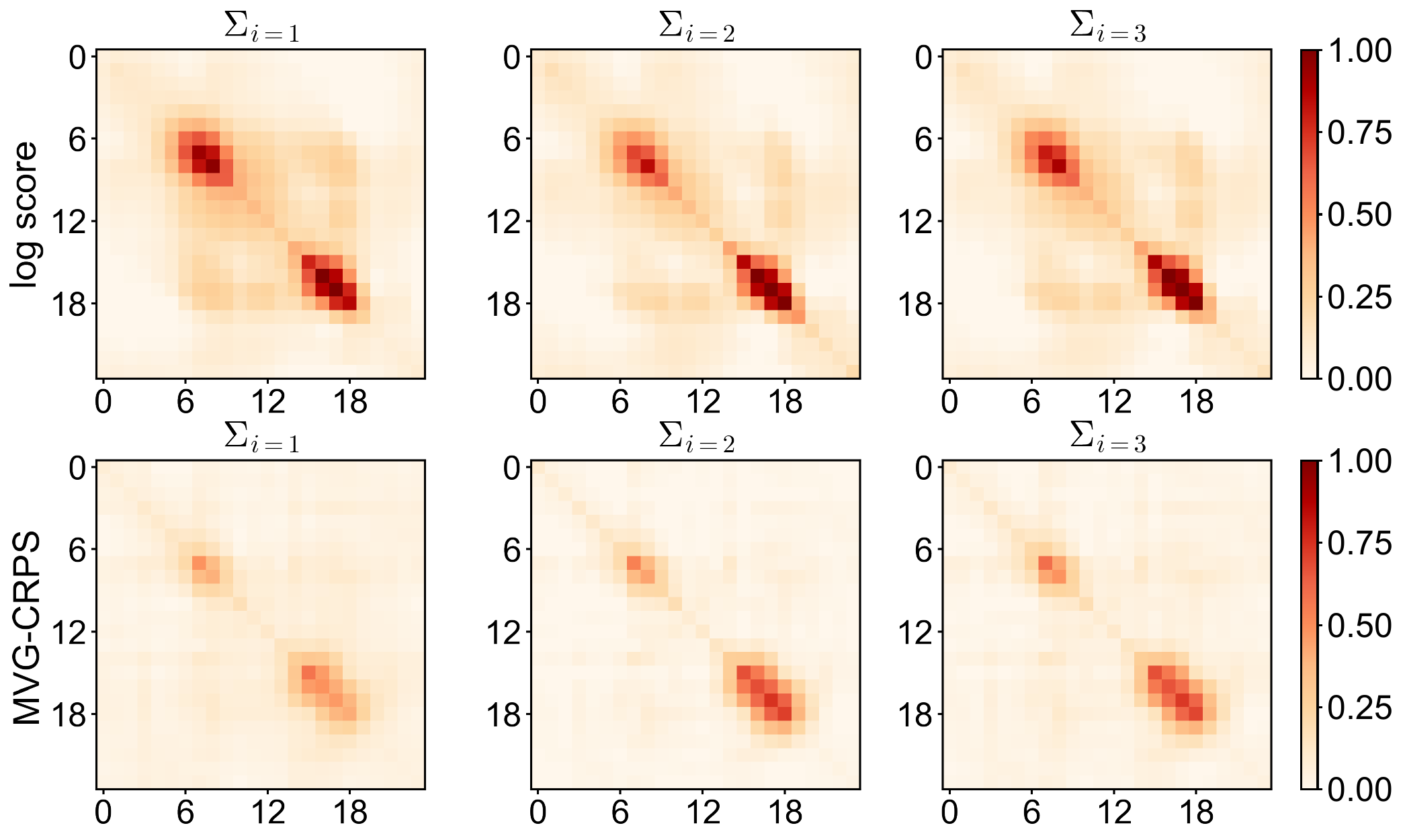}
  \caption{Comparison of output covariance matrices  \(\boldsymbol{\Sigma}_{i}\) from N-HiTS on the \(\mathtt{traffic}\) dataset. For visual clarity, covariance values are clipped between 0 and 1.0.}
\label{fig:cov_q}
\end{figure}

\begin{figure}[!t]
  \centering\includegraphics[width=0.46\textwidth]{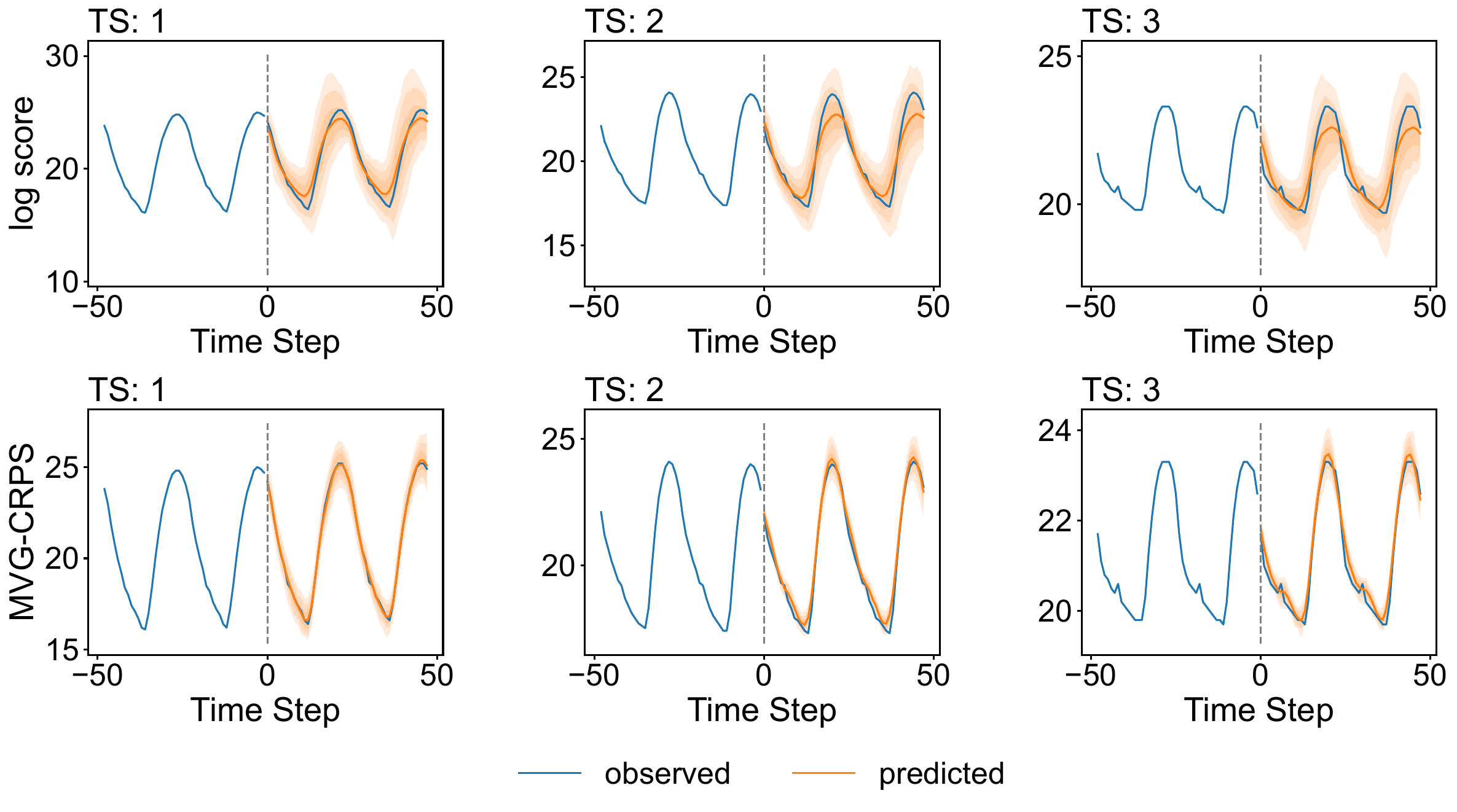}
  \caption{Comparison of probabilistic forecasts from N-HiTS on the $\mathtt{m4\_hourly}$ dataset.}
\label{fig:frcs_compare_q}
\end{figure}

The results for the univariate Seq2Seq forecasting task are presented in Fig.~\ref{fig:cov_q}. Consistent with the findings from the multivariate autoregressive task, the model trained with the log-score tends to produce higher variance and covariance values. Consequently, the log-score-trained model may exhibit increased uncertainty, which can degrade the reliability of its forecasts. Fig.~\ref{fig:cov_q} illustrates the covariance over the prediction horizon of one day in the hourly $\mathtt{traffic}$ dataset. The covariance matrices directly model the prediction uncertainty over time, influenced by both the prediction lead time and the time of day. Uncertainty is generally larger during rush hours and increases with longer lead times. Compared to log-score-trained model, the MVG-CRPS-trained model is less affected by tail values while still effectively capturing these temporal patterns. This demonstrates that MVG-CRPS maintains the ability to model relevant uncertainty trends without being unduly influenced by outliers.

Another example is shown in Fig.~\ref{fig:frcs_compare_q}, which depicts the probabilistic forecasts generated by the N-HiTS model on the $\mathtt{m4\_hourly}$ dataset. The MVG-CRPS-trained model produces narrower prediction intervals, suggesting greater confidence and improved calibration compared to the log-score-trained model. In contrast, the log-score-trained model generates overly wide intervals for simple time series with strong cyclical patterns. Additionally, the MVG-CRPS-trained model achieves higher accuracy for longer-term forecasts.

These qualitative observations highlight the benefits of MVG-CRPS for probabilistic forecasting. By reducing sensitivity to outliers and providing more stable covariance estimates, MVG-CRPS supports models in delivering accurate and reliable forecasts.





\section{Conclusion}

In this paper, we introduced MVG-CRPS, a loss function designed for multivariate Gaussian outputs in probabilistic forecasting. By addressing the sensitivity of the log-score to outliers and the inefficiency of the energy score, we leveraged the CRPS to improve robustness without sacrificing computational efficiency. Its closed-form expression based on neural network outputs allows seamless integration into deep learning models. Our experiments on real-world datasets show that MVG-CRPS provides improved accuracy and robustness in probabilistic forecasting tasks.

Furthermore, MVG-CRPS can be used for general tasks, including robust Gaussian process regression, by replacing the negative marginalized likelihood. While our focus has been on forecasting applications, the general nature of MVG-CRPS makes it applicable to a wide range of probabilistic regression problems involving multivariate Gaussian responses.

For future work, we plan to employ copulas to transform continuous distributions into Gaussian ones, extending our approach to non-Gaussian settings. We also aim to investigate improved parameterizations for covariance matrices to further enhance scalability. A limitation of our current approach is scalability with large batch sizes. Although a low-rank-plus-diagonal structure can be leveraged for the log-score through the Woodbury identity and the matrix-determinant lemma (when the batch size exceeds the factor dimension), MVG-CRPS does not benefit from this structure due to the SVD operation. A possible solution is to adopt an isotropic noise parameterization, i.e., $\boldsymbol{\Sigma}=\boldsymbol{L}\boldsymbol{L}^\top+\sigma^2 \mathbf{I}$, which enables more efficient computation of the SVD.




\section*{Impact Statement}

This paper presents work whose goal is to advance the field of 
Machine Learning. There are many potential societal consequences 
of our work, none which we feel must be specifically highlighted here.

\nocite{langley00}

\bibliography{refs}
\bibliographystyle{icml2025}

\newpage
\appendix
\onecolumn

\section*{Appendix}

\textbf{Table of Contents}\\
\noindent\makebox[\textwidth]{\rule{\textwidth}{0.4pt}}
\textbf{A Properties of MVG-CRPS} \dotfill \pageref{apx:proof}


\textbf{B Dataset Details} \dotfill \pageref{apx:data}

\textbf{C Experiment Details} \dotfill \pageref{apx:exp}

\hspace{0.5cm} C.1 Benchmark Models \dotfill \pageref{apx:exp_models}

\hspace{0.5cm} C.2 Hyperparameters \dotfill \pageref{apx:exp_hyparam} 

\hspace{0.5cm} C.3 Training Procedure \dotfill \pageref{apx:exp_train} 

\hspace{0.5cm} C.4 Covariance Parameterization \dotfill \pageref{apx:exp_cov} 

\hspace{0.5cm} C.5 SVD and Gradient Calculation \dotfill \pageref{apx:exp_svd} 

\hspace{0.5cm} C.6 Naive Baseline Description \dotfill \pageref{apx:exp_baseline} 

\textbf{D Additional Experiments} \dotfill \pageref{apx:additional_results}
\noindent\makebox[\textwidth]{\rule{\textwidth}{0.4pt}}

\section{Properties of MVG-CRPS}\label{apx:proof}



\begin{theorem}
Let $\mathbf{z}\sim \mathcal{N}\left(\boldsymbol{\mu}_p, \boldsymbol{\Sigma}_p\right)$ be a true $N$-variate Gaussian distribution where the covariance admits eigen-decomposition $\boldsymbol{\Sigma}_p=\boldsymbol{U}_p\boldsymbol{S}_p \boldsymbol{U}_p^{\top}$, with $\boldsymbol{S}_p=\operatorname{diag}\left(\boldsymbol{\lambda}_p\right)$ containing nonincreasing eigenvalues $\boldsymbol{\lambda}_p=\left[\lambda_1^p,\ldots,\lambda_N^p\right]^{\top}$ and $\boldsymbol{U}_p$ being the corresponding orthonormal matrix. Consider a predictive Gaussian distribution $\mathcal{N}\left(\boldsymbol{\mu}_q, \boldsymbol{\Sigma}_q\right)$, where covariance $\boldsymbol{\Sigma}_q$ admits the eigen-decomposition $\boldsymbol{\Sigma}_q=\boldsymbol{U}_q \boldsymbol{S}_q \boldsymbol{U}_q^{\top}$ with $\boldsymbol{S}_q=\operatorname{diag}\left(\boldsymbol{\lambda}_q\right)$. Define the transformed variable $\mathbf{v}=\boldsymbol{U}_q^{\top}\left(\mathbf{z}-\boldsymbol{\mu}_q\right)=\left[v_1,\ldots,v_N\right]^{\top}$. The proposed MVG-CRPS 
\begin{equation}\notag
\mathcal{L}\left(\mathcal{N}\left(\boldsymbol{\mu}_q, \boldsymbol{\Sigma}_q\right),\mathbf{z}\right) = \sum_{i=1}^N \operatorname{CRPS}\left(\Phi\left(0,\lambda_i^q\right), v_i\right)
\end{equation}
is proper and strictly proper for multivariate Gaussian distributions. 
\end{theorem}

\begin{proof}
Since $\mathbf{z}\sim \mathcal{N}\left(\boldsymbol{\mu}_p, \boldsymbol{\Sigma}_p\right)$, we have the transformed variable $\mathbf{v}\sim\mathcal{N}\left(\boldsymbol{\mu}_v,\boldsymbol{\Sigma}_v\right)$ with $\boldsymbol{\mu}_v=\boldsymbol{U}_q^{\top}(\boldsymbol{\mu}_p-\boldsymbol{\mu}_q)=\left[\nu_1,\ldots,\nu_N\right]^{\top}$ and $\boldsymbol{\Sigma}_v=\boldsymbol{U}_q^{\top} \boldsymbol{\Sigma}_p \boldsymbol{U}_q= \boldsymbol{U}_q^{\top}\boldsymbol{U}_p\boldsymbol{S}_p \boldsymbol{U}_p^{\top}\boldsymbol{U}_q=\boldsymbol{U}_v\boldsymbol{S}_p\boldsymbol{U}_v^{\top}$, where $\boldsymbol{U}_v=\boldsymbol{U}_q^{\top}\boldsymbol{U}_p$ is an orthonormal matrix. Thus, each $v_i$ has a marginal distribution $v_i\sim \mathcal{N}\left(\nu_i,\phi_i\right)$ with $\boldsymbol{\phi}=\operatorname{diag}(\boldsymbol{\Sigma}_v)=\operatorname{diag}(\boldsymbol{U}_v\boldsymbol{S}_p\boldsymbol{U}_v^{\top})=\left[\phi_1,\ldots,\phi_N\right]^{\top}$ for $i=1,\ldots,N$. Taking the expectation of $\mathcal{L}\left(\mathcal{N}\left(\boldsymbol{\mu}_q, \boldsymbol{\Sigma}_q\right),\mathbf{z}\right)$ under the true distribution, we have
\begin{equation}\label{equ:proof}
\begin{split}
\mathbb{E}_{\mathbf{z}\sim \mathcal{N}\left(\boldsymbol{\mu}_p, \boldsymbol{\Sigma}_p\right)}\left[ \mathcal{L}\left(\mathcal{N}\left(\boldsymbol{\mu}_q, \boldsymbol{\Sigma}_q\right),\mathbf{z}\right)\right] &= \sum_{i=1}^N \mathbb{E}_{v_i\sim\mathcal{N}\left(\nu_i,\phi_i\right)} \left[\operatorname{CRPS}\left(\Phi\left(0,\lambda_i^q\right), v_i\right)\right]\\
& \ge \sum_{i=1}^N \mathbb{E}_{v_i\sim \mathcal{N}\left(\nu_i,\phi_i\right)} \left[\operatorname{CRPS}\left(\Phi\left(\nu_i,\phi_i\right), v_i\right)\right] \\
   &=   \sum_{i=1}^N \mathbb{E}_{\eta_i\sim \mathcal{N}\left(0,\phi_i\right)} \left[\operatorname{CRPS}\left(\Phi\left(0,\phi_i\right), \eta_i\right)\right] \\
   & =  \mathbb{E}_{v\sim\mathcal{N}\left(0,1\right)}\left[\operatorname{CRPS}\left(\Phi, v\right)\right] \times  \sum_{i=1}^N \sqrt{\phi_i} \\
   & \ge \mathbb{E}_{v\sim\mathcal{N}\left(0,1\right)}\left[\operatorname{CRPS}\left(\Phi, v\right)\right] \times  \sum_{i=1}^N \sqrt{ \lambda_i^p} \\
   &= \mathbb{E}_{\mathbf{z}\sim \mathcal{N}\left(\boldsymbol{\mu}_p, \boldsymbol{\Sigma}_p\right)}\left[ \mathcal{L}\left(\mathcal{N}\left(\boldsymbol{\mu}_p, \boldsymbol{\Sigma}_p\right),\mathbf{z}\right)\right].
\end{split}
\end{equation}

The first inequality is a direct result of  CRPS being a strictly proper scoring rule for univariate Gaussian distributions. We now prove the second inequality.

Recall that  $\boldsymbol{\phi}=\operatorname{diag}(\boldsymbol{\Sigma}_v)$ and $\boldsymbol{\Sigma}_v=\boldsymbol{U}_q^{\top}\boldsymbol{\Sigma}_p\boldsymbol{U}_q$. Let $\boldsymbol{\phi}^*$ be the monotone nonincreasing rearrangement of $\boldsymbol{\phi}$. By the Schur-Horn theorem \cite{horn1954doubly}, the vector $\boldsymbol{\phi}^*$ is submajorized by the eigenvalues  $\boldsymbol{\lambda}_p$:
\begin{equation}\notag
    \sum_{i=1}^{k} \phi_i^* \leq \sum_{i=1}^{k} \lambda_i^p,
\end{equation}
for $k = 1,2,\dots,N-1$, and
\begin{equation}\notag
    \sum_{i=1}^N \phi_i^* = \sum_{i=1}^N \lambda_i^p.
\end{equation}
Since $f(x)=\sqrt{x}$ is a concave function, Karamata's majorization inequality yields
\begin{equation}\label{equ:karamata}
    \sum_{i=1}^N \sqrt{\lambda_i^p}\le\sum_{i=1}^N \sqrt{\phi_i^*}= \sum_{i=1}^N \sqrt{\phi_i},
\end{equation}
which proves the second inequality in Eq.~\eqref{equ:proof}.

Equality in Eq.~\eqref{equ:karamata} is obtained if, for every $i$, $\phi_i^*=\lambda_i^p$. By the Schur-Horn theorem, this forces $\boldsymbol{\Sigma}_v$ has to be a diagonal matrix (Theorem 4.3.45 in \citet{horn2012matrix}). Meanwhile, the CRPS inequality in Eq.~\eqref{equ:proof} is tight exactly when, for every $i$, $\nu_i=0$ and $\phi_i=\lambda_i^q$, implying that  
$\boldsymbol{U}_q^{\top} (\boldsymbol{\mu}_p - \boldsymbol{\mu}_q) = \boldsymbol{0}$ and $\operatorname{diag}(\boldsymbol{\Sigma}_v)=\operatorname{diag}(\boldsymbol{S_q})$. Since $\boldsymbol{\Sigma}_v$ is diagonal, we have $\boldsymbol{\Sigma}_v=\boldsymbol{U}_q^{\top}\boldsymbol{\Sigma}_p\boldsymbol{U}_q=\boldsymbol{S}_q$, hence $\boldsymbol{\Sigma}_p=\boldsymbol{\Sigma}_q$.
Therefore, all equalities hold if and only if \(\boldsymbol{\mu}_p = \boldsymbol{\mu}_q\) and \(\boldsymbol{\Sigma}_p = \boldsymbol{\Sigma}_q\). This confirms that the proposed scoring rule is proper and strictly proper for multivariate Gaussian distributions.
\end{proof}

\section{Dataset Details}\label{apx:data}

We conducted experiments on a diverse collection of real-world datasets sourced from GluonTS \citep{alexandrov2020gluonts}. These datasets are commonly used for benchmarking time series forecasting models, following their default configurations in GluonTS, which include granularity, prediction horizon ($Q$), and the number of rolling evaluations. For each dataset, we sequentially split the data into training, validation, and testing sets, ensuring that the temporal length of the validation set matched that of the testing set. The temporal length of the testing set was based on the prediction horizon and the required number of rolling evaluations. For example, the testing horizon for the $\mathtt{traffic}$ dataset is calculated as $24+7-1=30$ time steps. Consequently, the model generates 24-step predictions ($Q$) sequentially, with 7 distinct consecutive prediction start points, corresponding to 7 forecast instances. In our experiments, we aligned the conditioning range ($P$) with the prediction horizon ($Q$), consistent with the default setting in GluonTS (i.e., $P=Q$). Each dataset was normalized using the mean and standard deviation computed from each time series in the training set, and predictions were rescaled to their original values for evaluation. Table~\ref{tab:datasets} summarizes the statistics of all datasets.

\begin{table}[ht]
\small
  \caption{Dataset summary.}
  \label{tab:datasets}
  \centering
  \begin{tabular}{lccccc}
    \toprule
    Dataset     &   Granularity   &  \# of time series & \# of time steps & $Q$ & Rolling evaluation \\
    \midrule
    $\mathtt{elec\_au}$       & 30min     & 5    & 232,272 & 60 & 56 \\
    $\mathtt{cif\_2016}$      & monthly   & 72   & 120     & 12 & 1  \\
    $\mathtt{electricity}$    & hourly    & 370  & 5,857   & 24 & 7  \\
    $\mathtt{elec\_weekly}$   & weekly    & 321  & 156     & 8  & 3  \\
    $\mathtt{exchange\_rate}$ & workday   & 8    & 6,101   & 30 & 5  \\
    $\mathtt{kdd\_cup}$       & hourly    & 270  & 10,920  & 48 & 7  \\
    $\mathtt{m1\_yearly}$     & yearly    & 181  & 169     & 6  & 1  \\
    $\mathtt{m3\_yearly}$     & yearly    & 645  & 191     & 6  & 1  \\
    $\mathtt{nn5\_daily}$     & daily     & 111  & 791     & 56 & 5  \\
    $\mathtt{saugeenday}$     & daily     & 1    & 23,741  & 30 & 5  \\
    $\mathtt{sunspot}$        & daily     & 1    & 73,924  & 30 & 5  \\
    $\mathtt{tourism}$        & quarterly & 427  & 131     & 8  & 1  \\
    $\mathtt{traffic}$        & hourly    & 963  & 4,025   & 24 & 7  \\
    $\mathtt{covid}$          & daily     & 266  & 212     & 30 & 5  \\
    $\mathtt{elec\_hourly}$   & hourly    & 321  & 26,304  & 48 & 7  \\
    $\mathtt{m4\_hourly}$     & hourly    & 414  & 1,008   & 48 & 7  \\
    $\mathtt{pedestrian}$     & hourly    & 66   & 96,432  & 48 & 7  \\
    $\mathtt{taxi\_30min}$    & 30min     & 1214 & 1,637   & 24 & 56 \\
    $\mathtt{uber\_hourly}$   & hourly    & 262  & 8,343   & 24 & 7  \\
    $\mathtt{wiki}$           & daily     & 2000 & 792     & 30 & 5   \\
    \bottomrule
  \end{tabular}
\end{table}

\section{Experiment Details}\label{apx:exp}

\subsection{Benchmark Models}\label{apx:exp_models}
The input to benchmark models includes lagged time series values and covariates that encode time and series identification. The number of lagged values is determined by the granularity of each dataset. Specifically, we use lags of [1, 24, 168] for hourly data, [1, 7, 14] for daily data, and [1, 2, 4, 12, 24, 48] for data with sub-hourly granularity. For all other datasets, only lag-1 values are used.

For datasets with hourly or finer granularity, we include the hour of the day and day of the week. For daily datasets, only the day of the week is used. Each time series is uniquely identified by a numeric identifier. All features are encoded as single values; for example, the hour of the day takes values between [0, 23]. These features are concatenated with the model input at each time step to form the model input vector \(\mathbf{y}_{t}\) \citep{salinas2019high,zheng2024multivariate}.

Our method requires a state vector \(\mathbf{h}_{i,t}\) to generate the parameters for the predictive distribution. To achieve this, we employ different neural architectures: RNNs and Transformer decoders, both of which maintain autoregressive properties for the multivariate autoregressive forecasting task, and MLPs for the univariate Seq2Seq forecasting task. Specifically, we use the GPVar model \citep{salinas2019high} as our RNN benchmark, the GPT model \citep{radford2018improving} for the decoder-only Transformer, and the N-HiTS model \citep{challu2023nhits} for the MLPs. All models are trained to output \(\mathbf{h}_{i,t}\), which is used to parameterize the predictive distribution.

\subsection{Hyperparameters}\label{apx:exp_hyparam}
All model parameters are optimized using the Adam optimizer with \(l_2\) regularization set to 1e-8, and gradient clipping applied at 10.0. For all methods, we cap the total number of gradient updates at 10,000 and reduce the learning rate by a factor of 2 after 500 consecutive updates without improvement. Table~\ref{tab:hyparam} provides the hyperparameter values that remain fixed across all datasets. We did \textbf{NOT} tune the hyperparameters specifically to favor the proposed loss. Instead, we used the same hyperparameters as those in GPVar \citep{salinas2019high}, which were originally tuned for the log-score. Keeping the hyperparameters consistent across losses ensures that any observed improvements are attributable to the loss function itself, rather than differences in hyperparameter settings.

\begin{table}[htbp]
\small
\caption{Hyperparameters values.}\label{tab:hyparam}
\begin{center}
\begin{tabular}{l|c}
\toprule
Hyperparameter             & Value   \\
\midrule
learning rate              & 1e-3        \\
hidden size                 & 40              \\
$\mathtt{n\_layers}$ (RNN/Transformer decoder/MLP)          & 2                \\
$\mathtt{n\_heads}$ (Transformer)      & 2                 \\
rank (\(R\))                      & 10                \\
sampling dimension (\(B\))        & 20                         \\
dropout                    & 0.01                         \\
batch size                 & 16                        \\
\bottomrule
\end{tabular}
\end{center}
\end{table}

\subsection{Training Procedure}\label{apx:exp_train}

\textbf{Compute Resources} All models were trained in an Anaconda environment using one AMD Ryzen Threadripper PRO 5955WX CPU and four NVIDIA RTX A5000 GPUs, each with 24 GB of memory.

\textbf{Batch Size} Following the method used in GPVar \citep{salinas2019high}, we set the sample slice size to $B=20$ time series and used a batch size of 16. Since our data sampler processes one slice of time series at a time rather than sampling 16 slices simultaneously, we set $\mathtt{accumulate\_grad\_batches}$ to 16, effectively achieving a batch size of 16.

\textbf{Training Loop} During each epoch, the model is trained on up to 400 batches from the training set, followed by the computation of the $\mathtt{valid\_loss}$ on the validation set. Training is halted when one of the following conditions is met:
\begin{itemize}[nosep, noitemsep]
\item A total of 10,000 gradient updates has been reached,
\item No improvement in the validation set $\mathtt{valid\_loss}$ is observed for 10 consecutive epochs.
\end{itemize}
The final model is the one that achieves the lowest $\mathtt{valid\_loss}$ on the validation set.

\subsection{Covariance Parameterization}\label{apx:exp_cov}

The covariance matrix $\boldsymbol{\Sigma}_t$ is parameterized directly by the forecasting model. Specifically, it is constructed as:
$
\boldsymbol{\Sigma}_t = \boldsymbol{L}_t \boldsymbol{L}_t^\top + \text{diag}(\mathbf{d}_t),
$
where $\boldsymbol{L}_t$ is a low-rank matrix and $\mathbf{d}_t$ is a positive diagonal vector. This parameterization ensures that $\boldsymbol{\Sigma}_t$ remains positive semi-definite while being computationally efficient to learn. This parameterization is standard in probabilistic forecasting and allows the model to learn both the structure (through $\boldsymbol{L}_t$) and scale (through $\mathbf{d}_t$) of the covariance during training.

Without constraints, the MVG-CRPS loss could potentially be minimized by driving all eigenvalues of $\boldsymbol{\Sigma}_t$ to zero, resulting in a trivial solution. However, this is prevented through the following mechanisms:

\begin{itemize}
    \item The diagonal entries of the covariance matrix are parameterized as $d_{i,t} = \mathtt{softplus}(d_{i,t} + \mathtt{diag\_bias}) + \sigma_\text{min}^2$, where the \(\mathtt{softplus}\) function ensures that the diagonal entries are strictly positive, regardless of the raw input values, \(\mathtt{diag\_bias}\) is initialized to approximately $\mathtt{softplus\_inv}(\sigma_\text{init}^2)$, ensuring that the diagonal entries are initially close to $\sigma_\text{init}^2$. For instance, with $\sigma_\text{init} = 1.0$, the initial diagonal values start near 1.0. The addition of $\sigma_\text{min}^2$ provides a lower bound on the diagonal entries, ensuring that eigenvalues cannot approach zero.
    \item The low-rank component is parameterized as $\mathbf{L}_{i, t} = \frac{\mathbf{L}_{i, t}}{\sqrt{R}}$, where dividing by rank ensures that the low-rank term is well-scaled relative to the diagonal entries. This normalization prevents the low-rank component from dominating or becoming disproportionately small in the covariance matrix.
\end{itemize}

Moreover, the MVG-CRPS loss provides a balance between the calibration and sharpness of the forecasts:
\[
\mathbf{w}_{t} = \boldsymbol{S}_t^{-\frac{1}{2}} \mathbf{v}_{t} = \boldsymbol{S}_t^{-\frac{1}{2}} \boldsymbol{U}_t^{\top} \left( \mathbf{z}_{t} - \boldsymbol{\mu}_{t} \right),
\]
\[
\mathcal{L}= \sum_{t=1}^{T} \sum_{i=1}^N \sqrt{\lambda_{t}^i}\operatorname{CRPS}\left(\Phi,{w}_{i,t}\right).
\]
We observe that if the eigenvalues $\lambda_{t}^i$ in $\boldsymbol{S}_t$ approach zero, $w_{i,t}$ will be scaled very aggressively. This leads to inflated residuals $w_{i,t}$, which subsequently affect the CRPS computation. Since the CRPS metric integrates over the forecast distribution $F(y)$, penalizing deviations between $F(y)$ and the empirical step function $\mathbf{1}(y \geq w_{i,t})$, artificially large $w_{i,t}$ values (resulting from extreme eigenvalue scaling) will cause the CRPS term to increase significantly. This behavior reflects the importance of ensuring that eigenvalues $\lambda_{t}^i$ are well-regularized to prevent distortion in the forecast evaluation. By balancing the eigenvalue contributions, the MVG-CRPS ensures both stable calibration and sharpness in probabilistic forecasting.

\subsection{SVD and Gradient Calculation}\label{apx:exp_svd}

We perform SVD on $\boldsymbol{\Sigma}(\mathbf{h}_t)$ to obtain $\mathbf{U}_t$ and $\mathbf{S}_t$ (the eigenvectors and eigenvalues, respectively). These are required to compute the whitening transformation: $\mathbf{w}_t = \mathbf{S}_t^{-\frac{1}{2}} \mathbf{U}_t^\top (\mathbf{z}_t - \boldsymbol{\mu}_t).$ During training, gradients of $\mathcal{L}$ need to flow back through the whitened vecotr $\mathbf{w}_t$, the eigenvectors matrix $\mathbf{U}_t$, the eigenvalues matrix $\mathbf{S}_t$, and the covariance matrix $\boldsymbol{\Sigma}_t$. The gradient of $\mathcal{L}$ with respect to $\mathbf{w}_t$ is $\frac{\partial \mathcal{L}}{\partial \mathbf{w}_t}$. Gradients of $\mathbf{w}_t$ are propagated to the whitening transformation: $\mathbf{w}_t = \mathbf{S}_t^{-\frac{1}{2}} \mathbf{U}_t^\top (\mathbf{z}_t - \boldsymbol{\mu}_t)$, which involves: (1) gradients with respect to $\mathbf{U}_t$; (2) gradients with respect to $\mathbf{S}_t^{-\frac{1}{2}}$ (i.e., the square root and inverse of singular values); and (3) gradients with respect to $(\mathbf{z}_t - \boldsymbol{\mu}_t)$. Using PyTorch’s $\mathtt{torch.linalg.svd}$, we calculate the gradients of $\mathbf{U}_t$ and $\mathbf{S}_t$ via automatic differentiation. For the forward pass, the cost of SVD for $\boldsymbol{\Sigma}(\mathbf{h}_t) \in \mathbb{R}^{B \times B}$ is $O(B^3)$, where $B$ is the matrix dimension. For the backward pass, computing the gradients of $\mathbf{U}_t$ and $\mathbf{S}_t$ also incurs $O(B^3)$ computational cost. Memory usage scales as $O(B^2)$ for storing the covariance matrix and the singular value decomposition outputs ($\mathbf{U}_t, \mathbf{S}_t$). Additional memory is required for autograd intermediate values, scaling as $O(B^3)$. By leveraging PyTorch’s autograd system, we integrate the computation of $\mathbf{U}_t$, $\mathbf{S}_t$, and their gradients seamlessly into our end-to-end learning pipeline. This ensures that the whitening transformation and the loss function are fully differentiable, allowing the model parameters to be trained via gradient-based optimizers. The parameter $B$ also plays a crucial role in the scalability of our method. By leveraging the Gaussian assumption, we are able to train the model using a much smaller subset of time series at each step. Consequently, the size of the covariance matrix is reduced to $B \times B$, as opposed to $N \times N$, where $N$ represents the total number of time series in the dataset. This design ensures that the computational complexity of our method does not scale with $N$. Moreover, $B$ is kept relatively small in our implementation (e.g., $B=20$), making the approach computationally efficient.

\subsection{Naive Baseline Description}\label{apx:exp_baseline}

In this paper, we use Vector Autoregression (VAR) \citep{lutkepohl2005new} as a naive baseline model. The VAR(\(p\)) model is formulated as
\begin{equation}\label{var}
    \mathbf{z}_{t} = \mathbf{c} + \boldsymbol{A}_1\mathbf{z}_{t-1} + \dots + \boldsymbol{A}_p\mathbf{z}_{t-p} + \boldsymbol{\epsilon}_t, \quad \boldsymbol{\epsilon}_t \sim \mathcal{N}(\mathbf{0},\boldsymbol{\Sigma}_{\epsilon}),
\end{equation}
where \(A_i\) is an \(N \times N\) coefficient matrix, and \(\mathbf{c}\) is the intercept term. In our experiments, we employ a VAR model with a lag of 1 (VAR(1)). The parameters in Eq.~\eqref{var} are estimated using ordinary least squares (OLS), as described in \citet{lutkepohl2005new}. VAR models are not applied to datasets with insufficient time series in the testing set and are marked as ``N/A'' in this paper.

\section{Additional Results}\label{apx:additional_results}

In Table~\ref{tab:es} and Table~\ref{tab:es_ss}, we compare the energy score performance of models trained with different loss functions.

\begin{table*}[hbtp]
\setlength{\tabcolsep}{1pt}
\scriptsize
\caption{Comparison of energy score across different scoring rules in the multivariate autoregressive forecasting task. The best scores are in boldface. MVG-CRPS scores are underlined when they are not the best overall but exceed the log-score. Mean and standard deviation are reported from 10 runs of each model.}
\label{tab:es}
\begin{center}
\begin{tabular}{lccccccc}
\toprule
                          & VAR       & \multicolumn{3}{c}{GPVar}              & \multicolumn{3}{c}{Transformer} \\
\cmidrule(lr){3-8}
                          &             & log-score      & energy score       & MVG-CRPS           & log-score      & energy score       & MVG-CRPS   \\
\midrule
$\mathtt{elec\_au}$ (\(\times 10^3\))  & N/A   & 5.4013$\pm$0.0372   &   \textbf{3.9136$\pm$0.0177}     & \underline{4.1508$\pm$0.0283}         & 7.0039$\pm$0.0219      &  6.3135$\pm$0.0243       & \textbf{3.5217$\pm$0.0150}        \\
$\mathtt{cif\_2016}$ (\(\times 10^3\))                       & 125.6177$\pm$0.0000   & 4.2733$\pm$0.0218    &    4.9329$\pm$0.0161   & \textbf{4.1677$\pm$0.0198}         & 4.6316$\pm$0.0270      &  4.1063$\pm$0.0241       & \textbf{3.5559$\pm$0.0145}        \\
$\mathtt{elec}$ (\(\times 10^4\))               & 10.4788$\pm$0.0757   & 3.3124$\pm$0.0580   &    4.8317$\pm$0.0434  & \textbf{3.2435$\pm$0.0300}       & \textbf{3.4724$\pm$0.0229}  &   4.3757$\pm$0.0414
 & 3.9672$\pm$0.0374               \\
$\mathtt{elec\_weekly}$ (\(\times 10^7\))             & 2.2191$\pm$0.0308   & 2.5724$\pm$0.0799 &   \textbf{0.8948$\pm$0.02344}
 & \underline{1.4040$\pm$0.0887} & 1.5463$\pm$0.0582     &   \textbf{0.9338$\pm$0.0308}
 & \underline{0.9985$\pm$0.0360} \\
$\mathtt{exchange\_rate}$             & 0.1301$\pm$0.0002   & 0.3972$\pm$0.0074       &   0.1895$\pm$0.0034     & \textbf{0.1216$\pm$0.0013}             & 0.2136$\pm$0.0026    &  \textbf{0.1774$\pm$0.0026}    & \underline{0.2040$\pm$0.0045}            \\
$\mathtt{kdd\_cup}$ (\(\times 10^2\)) & N/A   & 4.7575$\pm$0.0186      &    4.3981$\pm$0.0164   & \textbf{4.0719$\pm$0.0180}           & 4.2809$\pm$0.0134       &  5.9466$\pm$0.0427        & \textbf{3.1788$\pm$0.0122}          \\
$\mathtt{m1\_yearly}$ (\(\times 10^4\))                      & N/A   & 7.3860$\pm$0.0789    &   7.7576$\pm$0.0335  & \textbf{6.1985$\pm$0.0505}       & 8.7079$\pm$0.1760     &  \textbf{5.7774$\pm$0.0755}     & \underline{7.5130$\pm$0.1784}     \\
$\mathtt{m3\_yearly}$ (\(\times 10^3\))                      & N/A   & 3.6113$\pm$0.0703    &    2.2147$\pm$0.0427   & \textbf{1.4775$\pm$0.0495}         & 3.1996$\pm$0.0995      &  4.0982$\pm$0.0621       & \textbf{2.4253$\pm$0.0914}        \\
$\mathtt{nn5\_daily}$ (\(\times 10^2\))     & 4.9419$\pm$0.0056   & \textbf{3.3001$\pm$0.0050} &   3.3004$\pm$0.0052
 & 3.3934$\pm$0.0045                    & 3.2546$\pm$0.0033        &  3.1622$\pm$0.0045       & \textbf{3.0996$\pm$0.0025}          \\
$\mathtt{saugeenday}$ (\(\times 10^2\))          & N/A   & 1.8098$\pm$0.0231 &   \textbf{1.7135$\pm$0.0150}
 & 1.9400$\pm$0.0208                    & \textbf{1.5780$\pm$0.0183}   &  1.5883$\pm$0.0108   & 1.8043$\pm$0.0204                   \\
$\mathtt{sunspot}$ (\(\times 10\))       & N/A   & 2.7737$\pm$0.1195      &   3.1658$\pm$0.0792     & \textbf{2.6195$\pm$0.1003}            & 5.4893$\pm$0.1132      &  \textbf{2.3153$\pm$0.0467}          & \underline{3.2663$\pm$0.0745}           \\
$\mathtt{tourism}$ (\(\times 10^5\))               & 3.5958$\pm$0.0354   & 6.1085$\pm$0.1132  &   5.6774$\pm$0.0493
  & \textbf{5.2111$\pm$0.0896}     & 5.0645$\pm$0.0526 &  \textbf{4.7502$\pm$0.0585}
 & 5.2702$\pm$0.0853             \\
$\mathtt{traffic\_nips}$                     & 3358.5004$\pm$10.7535   & 2.2924$\pm$0.0034  &   \textbf{2.1140$\pm$0.0023} & \underline{2.2916$\pm$0.0015}             & 2.2043$\pm$0.0012           &  2.2250$\pm$0.0018      & \textbf{2.2000$\pm$0.0018}       \\
\bottomrule
\end{tabular}
\end{center}
\end{table*}

\begin{table}[htbp]
\scriptsize
\caption{Comparison of energy score across different scoring rules in the univariate Seq2Seq forecasting task.}
\label{tab:es_ss}
\begin{center}
\begin{tabular}{lccc}
\toprule
                               & \multicolumn{3}{c}{N-HiTS}            \\
\cmidrule(lr){2-4}
                               & log-score        & energy score        & MVG-CRPS    \\
\midrule
$\mathtt{covid}$ (\(\times 10^5\))         & 2.1220$\pm$0.0304    & N/A                          & \textbf{0.9401$\pm$0.0186} \\
$\mathtt{elec\_hourly}$ (\(\times 10^5\))  & 0.9283$\pm$0.0161     & N/A                          & \textbf{0.9088$\pm$0.0079}  \\
$\mathtt{elec}$ (\(\times 10^5\))          & 0.2535$\pm$0.0018      & 0.3123$\pm$0.0019      & \textbf{0.2431$\pm$0.0020}  \\
$\mathtt{exchange\_rate}$ & 0.2876$\pm$0.0055            & 0.1272$\pm$0.0022            & \textbf{0.1240$\pm$0.0022}        \\
$\mathtt{m4\_hourly}$ (\(\times 10^4\))    & 0.2852$\pm$0.0026        & 0.2890$\pm$0.0029        & \textbf{0.2423$\pm$0.0027}    \\
$\mathtt{nn5\_daily}$ (\(\times 10^3\))    & 0.4170$\pm$0.0018          & \textbf{0.3272$\pm$0.0005} & \underline{0.3958$\pm$0.0021}         \\
$\mathtt{pedestrian}$ (\(\times 10^3\))    & 1.1571$\pm$0.0177        & 0.9746$\pm$0.0081          & \textbf{0.8337$\pm$0.0066}      \\
$\mathtt{saugeenday}$ (\(\times 10^2\))    & \textbf{1.6690$\pm$0.0391} & 1.7752$\pm$0.0216          & 1.7698$\pm$0.0129               \\
$\mathtt{taxi\_30min}$ (\(\times 10^2\))   & 6.9676$\pm$0.0045          & 6.7906$\pm$0.0058          & \textbf{5.6679$\pm$0.0004}      \\
$\mathtt{traffic}$        & 3.6810$\pm$0.0136            & 2.2524$\pm$0.0018            & \textbf{2.2200$\pm$0.0022}        \\
$\mathtt{uber\_hourly}$   & 6.3252$\pm$0.1785            & 5.4214$\pm$0.0326            & \textbf{4.2826$\pm$0.0320}        \\
$\mathtt{wiki}$ (\(\times 10^6\))          & 1.1535$\pm$0.0047   & 0.9352$\pm$0.0069     & \textbf{0.9338$\pm$0.0083}  \\
\bottomrule
\end{tabular}
\end{center}
\end{table}

\end{document}